\documentclass{article}

\PassOptionsToPackage{numbers, compress}{natbib}

\usepackage[preprint, nonatbib]{neurips_2024}

\usepackage[utf8]{inputenc} 
\usepackage[T1]{fontenc}    
\usepackage{hyperref}       
\usepackage{url}            
\usepackage{booktabs}       
\usepackage{amsfonts}       
\usepackage{nicefrac}       
\usepackage{microtype}      
\usepackage{xcolor}         

\usepackage{graphicx}
\usepackage{subcaption}
\usepackage{amsmath}
\usepackage{mathtools}
\usepackage{amssymb}
\usepackage{amsthm}
\usepackage{multirow,color}

\theoremstyle{plain}
\newtheorem{theorem}{Theorem}[section]

\newtheorem{lemma}[theorem]{Lemma}

\theoremstyle{definition}
\newtheorem{definition}[theorem]{Definition}
\newtheorem{assumption}[theorem]{Assumption}
\theoremstyle{remark}

\title{OwMatch: Conditional Self-Labeling with Consistency for Open-World Semi-Supervised Learning}

\author{
  Shengjie Niu$^{1*}$, Lifan Lin$^{2}$\thanks{The first two authors contributed equally to this work.}, Jian Huang$^{1}$, Chao Wang$^{2}$\thanks{Corresponding author.}\\\
  $^1$Hong Kong Polytechnic University, $^2$Southern University of Science and Technology \\
\texttt{shengjie.niu@connect.polyu.hk, 
12012816@mail.sustech.edu.cn, }\\
\texttt{j.huang@polyu.edu.hk, wangc6@sustech.edu.cn}\\ 
}

\begin{document}

\maketitle

\begin{abstract}
Semi-supervised learning (SSL) offers a robust framework for harnessing the potential of unannotated data. Traditionally, SSL mandates that all classes possess labeled instances. However, the emergence of open-world SSL (OwSSL) introduces a more practical challenge, wherein unlabeled data may encompass samples from unseen classes. This scenario leads to the misclassification of unseen classes as known ones, consequently undermining classification accuracy. To overcome this challenge, this study revisits two methodologies from self-supervised and semi-supervised learning, self-labeling and consistency, tailoring them to address the OwSSL problem. Specifically, we propose an effective framework called {\it OwMatch}, combining conditional self-labeling and open-world hierarchical thresholding.  Theoretically, we analyze the estimation of class distribution on unlabeled data through rigorous statistical analysis, thus demonstrating that OwMatch can ensure the unbiasedness of the self-label assignment estimator with reliability. 
Comprehensive empirical analyses demonstrate that our method yields substantial performance enhancements across both known and unknown classes in comparison to previous studies. Code is available at \href{https://github.com/niusj03/OwMatch}{https://github.com/niusj03/OwMatch}.

\end{abstract}

\section{Introduction}
\label{sec:intro}
Deep learning has made remarkable success in various tasks by leveraging substantial labeled training data \cite{resnet, 8237584, goodfellow2016deep}. However, the costly and time-consuming labeling process limits their application in practical scenarios. Semi-supervised learning (SSL) significantly reduces the dependency on labeled data by exploring the inherent structure of unlabeled data~\cite{guoSafeDeepSemiSupervised2020}.
Despite promising results, SSL methods assume a closed-world scenario where, though limited, all classes possess labeled instances. This assumption may be violated due to difficulties in data collection, such as in medical diagnostics, where it is common to encounter new symptoms or fail to annotate due to technical constraints. As a result, only a subset of the categories can be precisely labeled during the annotation process.
Recently, numerous studies have sought to identify such novel classes effectively. Open-world SSL (OwSSL) is innovative in promoting dual objectives: classifying instances of seen classes and discovering instances of novel classes \cite{caoOpenWorldSemiSupervisedLearning2022}.

A notable challenge in OwSSL is the \emph{confirmation bias} of model: model tends to predict instances as seen classes owing to the lack of ground-truth supervision of novel-class instances. To eliminate this bias, existing works utilize unsupervised clustering methods, including contrastive loss and binary cross-entropy (BCE) loss, to group pairs identified by similarity metrics~\cite{caoOpenWorldSemiSupervisedLearning2022, guoRobustSemiSupervisedLearning2022}.
Among these unsupervised techniques, self-labeling \cite{asanoSelflabellingSimultaneousClustering2020,caronUnsupervisedLearningVisual2020} has shown remarkable success, which involves assigning self-labels to unlabeled data, with the generation of high-quality self-labels being the key factor. Previous studies utilize optimal transport to align the self-labels for unlabeled data with a given distribution. However, this self-label generation fully relies on the accurate prior distribution and lack of consideration of the supervision of labeled data. In TRSSL \cite{rizve2022realistic}, the unlabeled data are assigned with a soft self-label based on the inaccurate class distribution, which raises a biased estimation. Moreover, the confirmation bias still exists even if we use the ground-truth distribution to align the unlabeled data in the same process. In addition to confirmation bias, a new issue called \emph{clustering misalignment} arises when self-labeling depends solely on unlabeled data: without proper guidance, the self-labeling process may adopt varying criteria for clustering. For example, it might cluster data based on superficial features like color rather than high-level semantic information. This misalignment can lead to results that deviate from expected outcomes and even contradict the classification criteria established by labeled data.

\begin{figure*}[t]
    \begin{subfigure}{0.5\textwidth}
        \centering
        \includegraphics[width=0.98\textwidth]{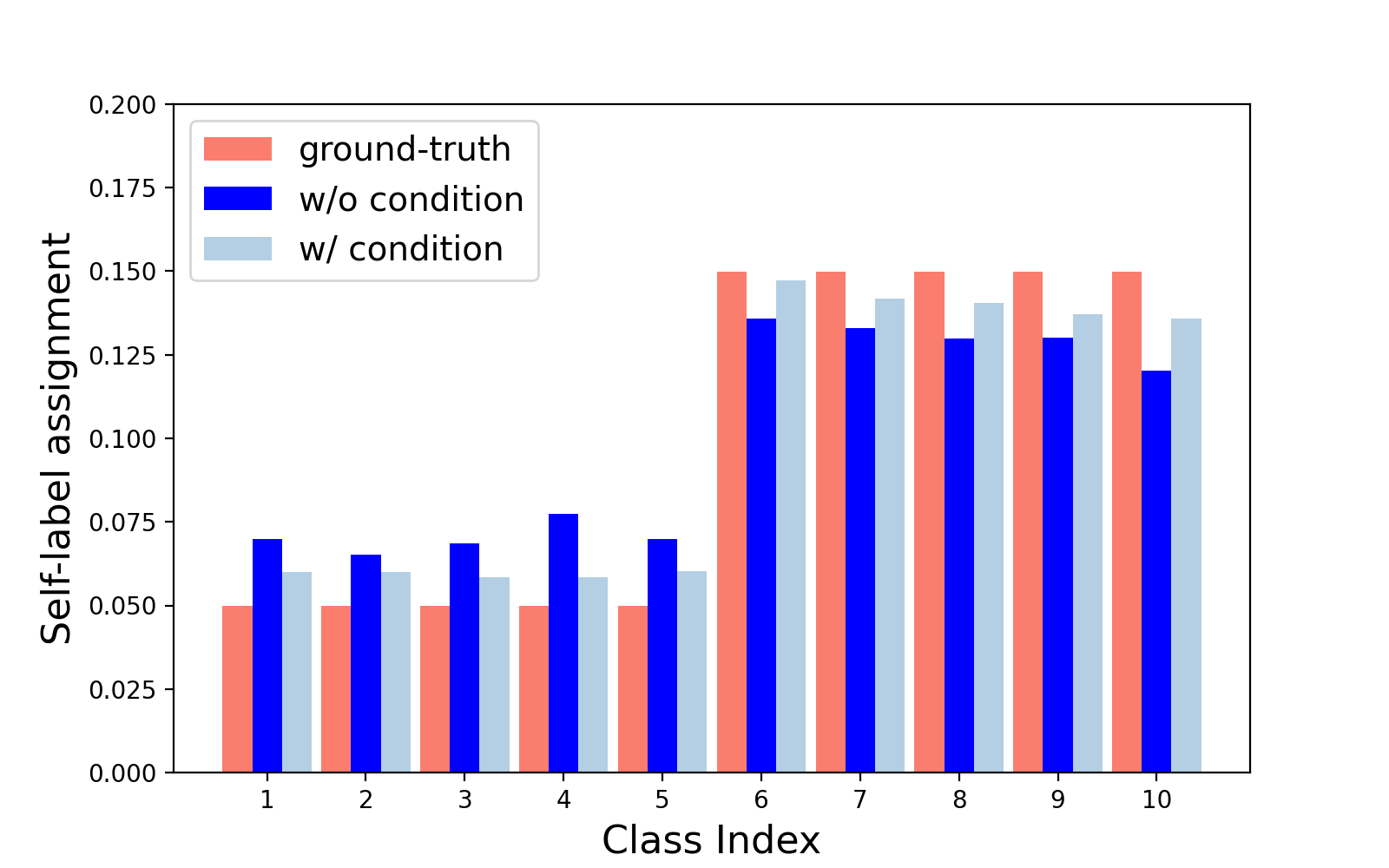}
        \caption{}
        \label{fig:condit}
    \end{subfigure}
    \begin{subfigure}{0.5\textwidth}
        \centering
        \includegraphics[width=0.98\textwidth]{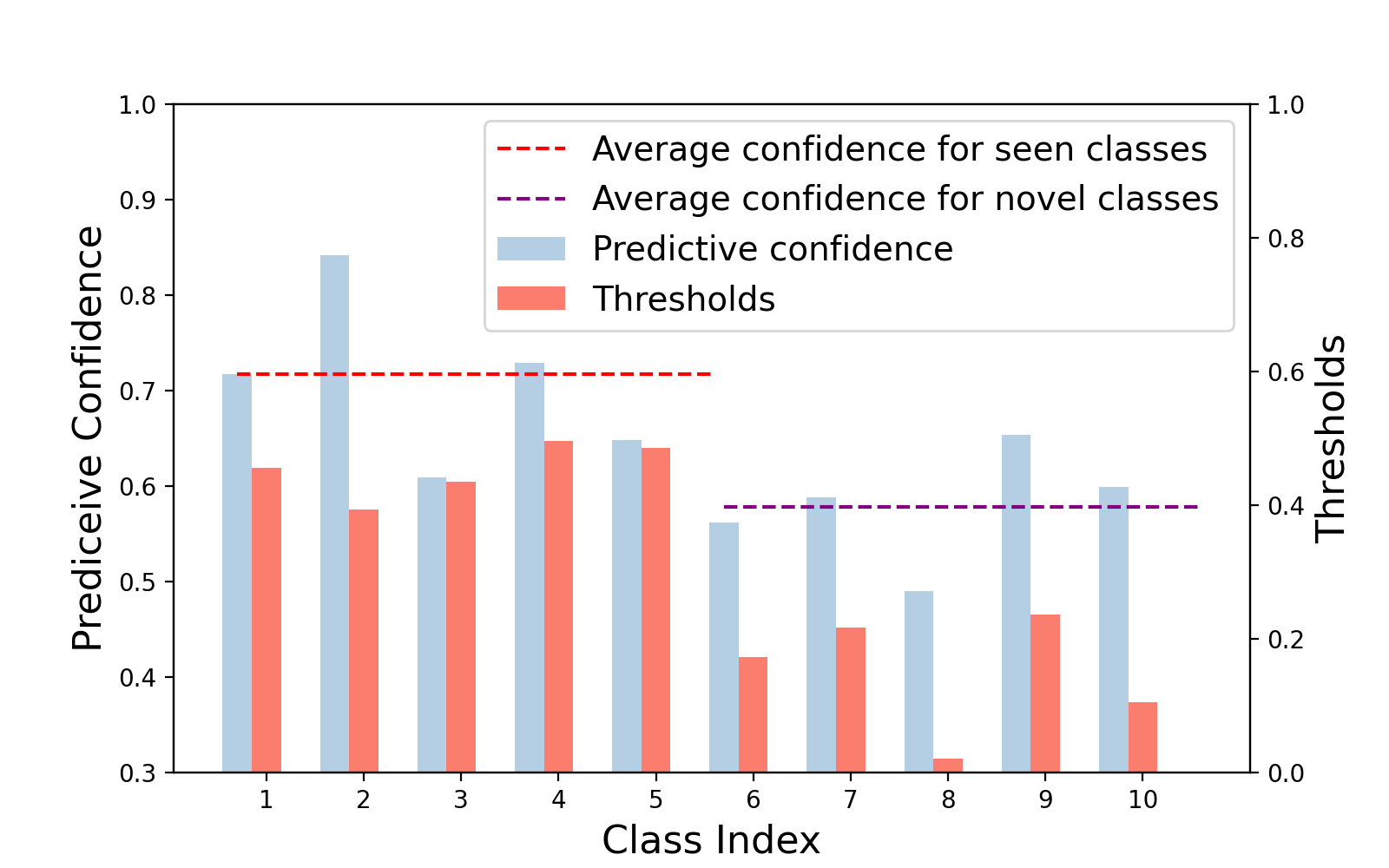}
        \caption{} 
        \label{fig:thres}
    \end{subfigure}
    \caption{Experimental results on the OwSSL problem. (a) Self-label assignment of seen classes (1-5) and novel classes (6-10) with or without conditional component in self-labeling. (b) Predictive confidence and hierarchical threshold for each class. }
    \vspace{-1em}
\end{figure*}

Consequently, we propose a new self-labeling scheme, conditional self-labeling, designed to address the challenges of OwSSL, particularly targeting issues related to confirmation bias and misalignment. This scheme limits self-labels for each class and incorporates labeled data to generate debiased and informative self-label assignments for all training data, further mitigating the confirmation bias as shown in Figure~\ref{fig:condit}. 
Additionally, as illustrated in Figure~\ref{fig:thres}, seen classes typically exhibit higher predictive confidence, while novel clusters demonstrate variability in their internal learning progresses. The disparities in learning paces between seen and novel classes, coupled with their distinct behaviors, necessitate the selection of appropriate thresholds to facilitate cluster learning. To address these challenges and ensure a balanced learning process across classes, we propose a hierarchical thresholding scheme.

We demonstrate our contributions as follows: \textbf{1)} We introduce a novel conditional self-labeling method to incorporate labeled data into the clustering process, reducing confirmation bias and misalignment. \textbf{2)} We design a hierarchical thresholding strategy that balances learning difficulties across different classes, helping unstable clusters gradually form. \textbf{3)} Our theoretical analysis rigorously discusses the unbiasedness and reliability of conditional self-labeling estimator from population-level statistics. To the best of our knowledge, this is the first work proposing the expectation of chi-square statistics (ECS) to evaluate the reliability of self-label assignment estimation.
\textbf{4)} 
We conduct extensive experiments on various datasets, demonstrating the effectiveness of our approach, OwMatch, through detailed comparisons. On CIFAR-10, OwMatch significantly outperforms FixMatch \cite{sohnFixMatchSimplifyingSemiSupervised2020} by up to 47.3\% in all-class accuracy, while on CIFAR-100, it enhances TRSSL \cite{rizve2022realistic}, the state-of-the-art model in OwSSL, by up to 14.6\% in novel-class and 7.2\% in all-class accuracy.

\section{Related work}
\paragraph{Traditional semi-supervised learning (SSL).}
Traditional SSL assumes that labeled and unlabeled data share an identical distribution. Extensive researches on SSL have spanned a considerable duration. The commonly employed strategies in SSL consist of entropy minimization \cite{pseudo_label, entropy_mini}, consistency regularization \cite{xieUnsupervisedDataAugmentation2020,8417973} and holistic methods \cite{berthelotMixMatchHolisticApproach2019, berthelotReMixMatchSemiSupervisedLearning2020, sohnFixMatchSimplifyingSemiSupervised2020}. The latest progresses in SSL include adaptive thresholding strategies \cite{wangFreeMatchSelfadaptiveThresholding2023, xuDashSemiSupervisedLearning2021, zhangFlexMatchBoostingSemiSupervised2022}, which enhance model performance by accounting for varying difficulties and learning conditions across classes, alongside other innovative techniques \cite{hanUnsupervisedSemanticAggregation2020, wangUnsupervisedSelectiveLabeling2022, assranSemiSupervisedLearningVisual2021} that employ self-SL approaches to facilitate extracting the semantic information from unlabeled data. However, traditional SSL algorithms typically struggle to tackle the open-world problem in the presence of novel-class instances within unlabeled data.

\paragraph{Open-set semi-supervised learning (OSSL). }
OSSL expands the traditional SSL boundaries by allowing novel-class instances or outliers within unlabeled data. A variety of OSSL approaches have emerged in recent years \cite{yuMultiTaskCurriculumFramework2020,guoSafeDeepSemiSupervised2020,saitoOpenMatchOpensetConsistency2021, chenSemiSupervisedLearningClass2020, T2T}. A common solution among these methods is the optimization of the SSL objective exclusively for unlabeled samples deemed inliers. For instance, MTC \cite{yuMultiTaskCurriculumFramework2020} optimizes the network and estimates the anomaly score of unlabeled data alternately. OpenMatch \cite{saitoOpenMatchOpensetConsistency2021} and T2T~\cite{T2T} train one-vs-all (OVA) classifiers for each known class to detect outliers. Subsequently, standard SSL objective \cite{sohnFixMatchSimplifyingSemiSupervised2020} are applied to the remaining training data, excluding detected outliers. Furthermore, DS3L \cite{guoSafeDeepSemiSupervised2020} leverages a bi-level optimization technique to train a weighting function, which mitigates the passive impact of out-of-distribution (OOD) samples. Nonetheless, these approaches are designed for classifying seen classes, thereby failing to learn from the novel class instances.

\paragraph{Open-world semi-supervised learning (OwSSL). }
OwSSL \cite{caoOpenWorldSemiSupervisedLearning2022} has been proposed to address a practical challenge: enabling the model to effectively cluster novel-class instances while maintaining classification robustness on seen classes.
One predominant research direction in this under-explored domain is BCE-based methods, including ORCA \cite{caoOpenWorldSemiSupervisedLearning2022} and NACH \cite{guoRobustSemiSupervisedLearning2022}. Additionally, there exist methods to discover novel classes by employing various other clustering techniques: OpenLDN \cite{rizveOpenLDNLearningDiscover2022} employs bi-level optimization to train a pairwise similarity prediction network, which provides a supervisory signal to the similarity of all pairs; TRSSL \cite{rizve2022realistic} converts clustering into the self-labeling problem and applies Sinkhorn-Knopp algorithm to optimize self-label assignments. One subsequently proposed Generalized Category Discovery (GCD) setting is similar to the OwSSL \cite{gcd2022,gpc}, with detailed discussion is provided in Section \ref{sec:ablation}.

\section{Methodology}

\paragraph{Problem setup.}
Given training data consisting of labeled data $\mathcal{D}_l=\{(\mathbf{x}^{(i)},\mathbf{y}_{\rm gt}^{(i)}) \}_{i=1}^{N^l}$ and unlabeled data $\mathcal{D}_u = \{\mathbf{x}^{(i)} \}_{i=N^l+1}^{N^l+N^u}$, where $N= N^l+N^u$ and $N^u \gg N^l$. 
Here $\mathbf{x}^{(i)} \in \mathbb{R}^d $ is the $i$-th instance with one-hot vector $\mathbf{y}_{\rm gt}^{(i)}\in \{0, 1\}^K$ as the corresponding label , where $K$ is the number of all classes. We denote the set of classes in $\mathcal{D}_l$ as $\mathcal{C}_l$ and the set of classes in $\mathcal{D}_u$ as $\mathcal{C}_u$. Previous traditional SSL studies assume $\mathcal{C}_l = \mathcal{C}_u$. Here for OwSSL, we assume $\mathcal{C}_l \neq \mathcal{C}_u $ and $\mathcal{C}_u \setminus \mathcal{C}_l \neq \emptyset$. Denote $\mathcal{C}_s = \mathcal{C}_l \cap \mathcal{C}_u $ as a set of seen classes, $\mathcal{C}_n = \mathcal{C}_u \setminus \mathcal{C}_l$ as a set of novel classes, and $\mathcal{C} = \mathcal{C}_l\cup \mathcal{C}_u$ as a set of all considered classes. The desired OwSSL model is required to assign instances to either a previously seen class $c\in\mathcal{C}_s$, or a novel class $c\in \mathcal{C}_n$.

For labeled dataset $\mathcal{D}_l$, standard supervised objective is employed as shown in Equation~\ref{eq:final_objective}.
Additionally, OwMatch primarily incorporates two objectives: a) clustering objective, which leverages conditional self-labeling to refine the self-label assignment with the assistance of supervision; b) confidence objective, which applies consistency loss with open-world hierarchical thresholding strategy to enhance predictive confidence and balance the different learning difficulties across all classes. We will elaborate on them respectively in Section~\ref{subsec:Con-SL} and \ref{subsec:thresholding}.

\begin{figure*}[t]
    \begin{minipage}{0.64\textwidth}
        \centering
        \includegraphics[width=0.98\textwidth]{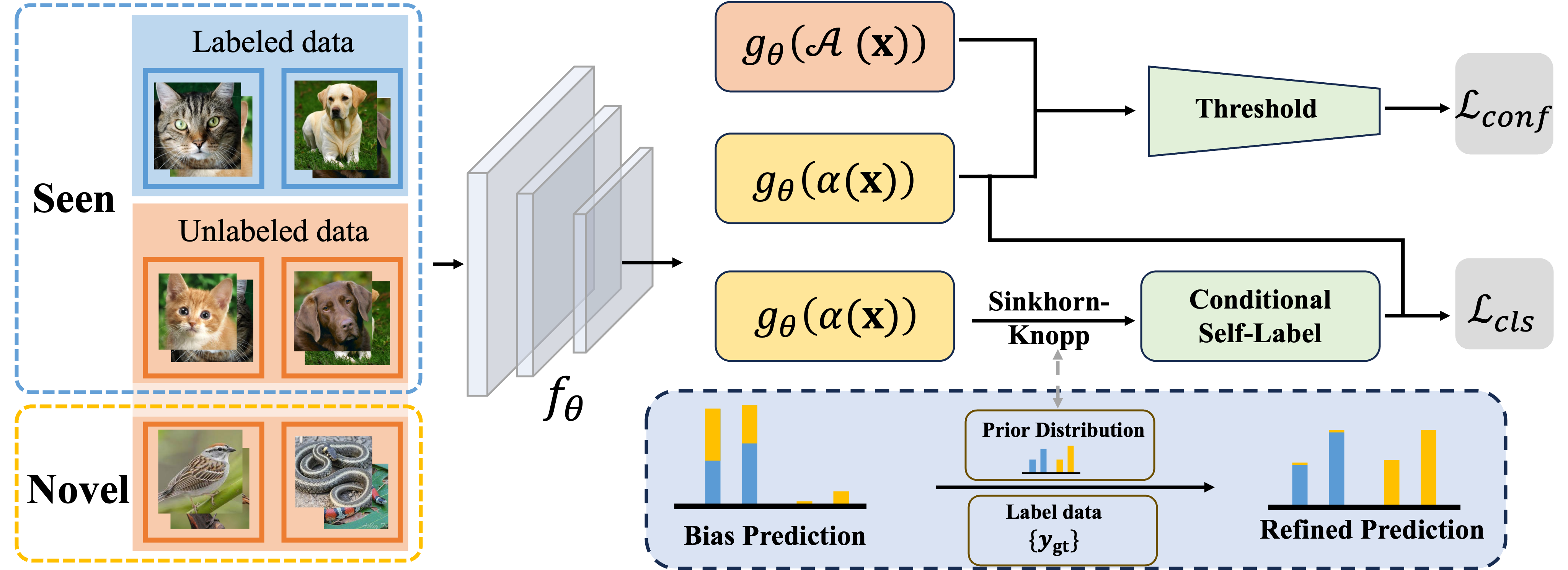}
        \caption{Overview of the OwMatch framework. }
        \label{fig:pipline}
    \end{minipage}\qquad
    \begin{minipage}{0.32\textwidth}
        \centering
        \includegraphics[width=0.9\textwidth]{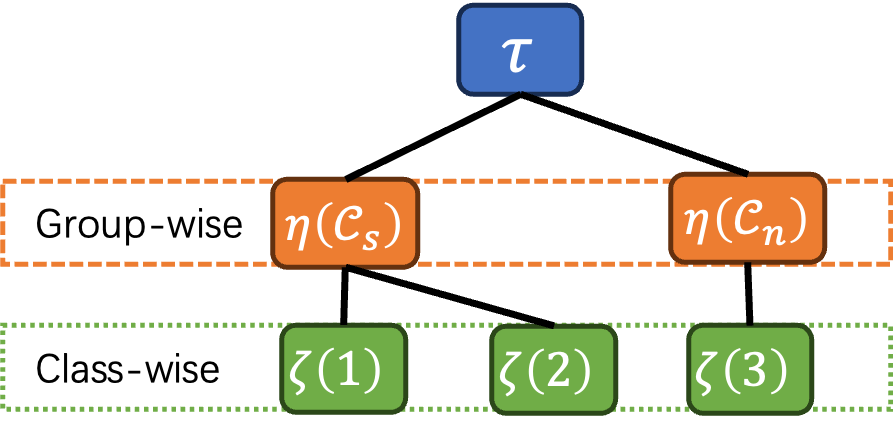}
        \caption{Illustration on the  hierarchical thresholding  
scheme. } 
        \label{fig:threshold}
    \end{minipage}
\end{figure*}

\subsection{Conditional self-labeling}
\label{subsec:Con-SL}

To effectively cluster the novel class instances, the self-labeling scheme \cite{asanoSelflabellingSimultaneousClustering2020} has been considered in OwSSL. Formally, consider a deep neural network (encoder) $f_{\theta}$ mapping input data $\mathbf{x}$ to representation $\mathbf{z}\in \mathbb{R}^D$, the representation is followed by a classification head $h: \mathbb{R}^D \rightarrow \mathbb{R}^{K}$, usually consisting of a single linear layer, converting the feature vectors into a vector of class scores. 
Denote $g_\theta = \sigma \circ h \circ f_\theta$ as a probability function, where $\sigma$ refers to the SoftMax function. Moreover, denote $\mathbf{q}^{(i)}\in\mathbb{R}^{K}$ as the soft self-label for $\mathbf{x}^{(i)}$, and set $\mathbf{Q} = [\mathbf{q}^{(1)}, \mathbf{q}^{(2)}, \dots, \mathbf{q}^{(N)}]\in \mathbb{R}^{K \times N}$ as the self-label assignment for $\{\mathbf{x}^{(i)}\}_{i=1}^N$. Asano et al. \cite{asanoSelflabellingSimultaneousClustering2020} utilize a constraint of desired partition of $\mathbf{Q}$ to construct the transportation polytope:
\begin{equation}
\label{eq:res_uncon}
     \mathcal{Q}_1:=\{\mathbf{Q} \in \mathbb{R}^{K \times N }_+ | \mathbf{Q} \mathbf{1}_N =N \boldsymbol{\mathcal{P}}, \mathbf{Q}^T \mathbf{1}_K=\mathbf{1}_{N} \},
\end{equation}
where $\mathbf{1}_{v}$ is the $v$-dimensional vector of all ones, $\boldsymbol{\mathcal{P}}$ denotes the desired class distribution. On the other hand, we can obtain a probability output through $\mathbf{p}^{(i)}= g_\theta(\alpha (\mathbf{x}^{(i)}))$, where $\alpha(\cdot)$ refers to a specific weak augmentation, and denote $\mathbf{P} = [\mathbf{p}^{(1)}, \mathbf{p}^{(2)}, \dots, \mathbf{p}^{(N)}]$ as the matrix of probability outputs. This self-label assignment generation can be understood as solving an optimal transport problem \cite{asanoSelflabellingSimultaneousClustering2020}. It minimizes the cross-entropy loss and aligns the training data with the desired class distribution: 
\begin{equation}
    \min_{\mathbf{Q}\in \mathcal{Q}_1} \operatorname{Tr}(\mathbf{Q} \log (\mathbf{P}^T)),  
\end{equation}
where $\operatorname{Tr}(\cdot)$ is the trace of a given matrix. 
Obviously, clustering through self-labeling primarily relies on the quality of generated self-label assignments. However, optimizing self-label assignments through unsupervised self-labeling is unreliable owing to the lack of supervision. TRSSL \cite{rizve2022realistic} utilizes the above-unsupervised technique to optimize self-label assignment merely on unlabeled data with the uniform class distribution. Despite prominent results, this unconditional self-labeling process \cite{rizve2022realistic} has a notable flaw: it constructs transportation polytope based on an inaccurate class distribution.

Moreover, we consider conducting self-labeling across all training data and constructing a transportation polytope with a precise class distribution.
To mitigate the confirmation bias, we propose a conditional self-labeling method to refine the self-label assignment under partial supervision. Specifically, we exploit the ground-truth labels from the labeled dataset and introduce another constraint $\mathcal{Q}_2$: 
\begin{equation}
    \mathcal{Q}_2 := \{\mathbf{Q} \in \mathbb{R}^{K \times N }_+ |  \mathbf{q}^{(i)} = \mathbf{y}^{(i)}_{\rm gt}, i = 1, \dots, N^l \}. 
    \label{eq:res_con}
\end{equation}
Now, the conditional self-label assignment generation with the above two constraints can be formulated as:
\begin{equation}
\label{eq:skmodel}
        \min_{\mathbf{Q}\in \mathcal{Q}_1\cap \mathcal{Q}_2 } \operatorname{Tr}(\mathbf{Q} \log (\mathbf{P}^T)) +\epsilon E(\mathbf{Q}),  
\end{equation}
where $E(\cdot)$ is the entropy function, $\epsilon$ is a hyper-parameter controlling the smoothness of $\mathbf{Q}$. We adopt fast version \cite{cuturi2013sinkhorn} of \textit{Sinkhorn-Knopp algorithm} to optimize Equation~\ref{eq:skmodel} efficiently and denote the optimal solution as $\tilde{\mathbf{Q}}=[\tilde{\mathbf{q}}^{(1)}, \tilde{\mathbf{q}}^{(2)}, \dots, \tilde{\mathbf{q}}^{(N)}]$.
Empirically, conditional self-labeling significantly alleviates the confirmation bias, resulting in self-label assignments that are much closer to the expected distribution, as shown in Figure~\ref{fig:condit}. Further theoretical analysis regarding estimators from unconditional and conditional self-labeling is provided in Section~\ref{sec:theoretical}. Then, the clustering objective has the form of: $\mathcal{L}_{cls} = \frac{1}{N}\sum_{i=1}^{N} H(\tilde{\mathbf{q}}^{(i)}, \mathbf{p}^{(i)}).$

\subsection{Open-world hierarchical thresholding}
\label{subsec:thresholding}
Beyond the clustering objective, prompting the predictive confidence has proven effective for classification. A similar goal arises in traditional SSL, wherein entropy minimization is employed to encourage low entropy (i.e., high confidence) in the prediction. FixMatch~\cite{sohnFixMatchSimplifyingSemiSupervised2020} leverages both consistency and pseudo-labeling to achieve exceptional performance with the following regularization:
$\sum_{i = 1}^N \mathbb{I}(\max(\mathbf{p}^{(i)})\geq \tau) H(\hat{\mathbf{p}}^{(i)}, g_\theta(\mathcal{A}(\mathbf{x}^{(i)}))), $
where $\hat{\mathbf{p}}^{(i)} := \arg \max(\mathbf{p}^{(i)})$ is predictive one-hot pseudo-label, with the $\hat{p}^{(i)}$-th element set to 1. $\alpha$ and $\mathcal{A}$ represent weak and strong augmentation respectively. Here, $\tau$ is a scalar hyperparameter denoting the threshold above which we retain a pseudo-label. 
The effectiveness of the aforementioned regularization depends on accurate and sufficient pseudo-labels, which are directly influenced by the thresholding scheme. Under the close-word assumption, extensive efforts \cite{zhangFlexMatchBoostingSemiSupervised2022,xuDashSemiSupervisedLearning2021,wangFreeMatchSelfadaptiveThresholding2023} have been devoted to devising thresholding techniques based on the idea of balancing learning pace across classes with varying learning difficulties. However, these techniques do not fit with the open-world scenario due to a critical challenge: the learning pace of novel classes tends to be much slower~\cite{caoOpenWorldSemiSupervisedLearning2022}. 
The predictive confidence of these two groups does not share the same behavior, as shown in Figure \ref{fig:thres}.

We introduce an open-world hierarchical thresholding scheme to balance this inconsistent learning pace at the group level, leveraging these well-defined thresholds to retain high-quality and adequate pseudo-labels for learning. As shown in Figure~\ref{fig:threshold}, this scheme first estimates the learning conditions of the two groups and then hierarchically modulates the thresholds in a class-specific fashion within each group.

First, we split the dataset into seen ($\mathcal{C}_s$) and novel ($\mathcal{C}_n$) groups based on the pseudo-label and estimate their overall learning condition by predictive confidence. Motivated by FreeMatch \cite{wangFreeMatchSelfadaptiveThresholding2023},    
we define the group-wise learning status for a set of classes $\mathcal{C}_i= \mathcal{C}_s$ or $\mathcal{C}_n$ as
\begin{equation}
        \eta(\mathcal{C}_i) = \frac{1}{N_{\mathcal{C}_i}}\sum_{i=1}^N \max (\mathbf{p}^{(i)})\mathbb{I}({\hat{p}^{(i)}\in \mathcal{C}_i)}, \ \mathcal{C}_i = \mathcal{C}_s \ \text{or} \ \mathcal{C}_n, 
\end{equation}
where $N_{\mathcal{C}_i}=\sum_{i=1}^N \mathbb{I}({\hat{p}^{(i)}\in \mathcal{C}_i})$ denotes the number of samples whose predictive pseudo-labels belong to the group $\mathcal{C}_i$. Similarly, the class-wise learning conditions can be defined as 
\begin{equation}
    \zeta_{c} = \frac{1}{N_c}\sum_{i=1}^N \max (\mathbf{p}^{(i)})\mathbb{I}({\hat{p}^{(i)}=c}), \ 
     c = 1, \dots, K,
\end{equation}
where $N_{c}=\sum_{i=1}^N \mathbb{I}({\hat{p}^{(i)}= c})$ denotes the number of samples whose predicted labels belong to the $c$-th class. In practice, we utilize the exponential moving average (EMA) to update at each iteration.
Then, we merge these two learning statuses and obtain the open-world hierarchical threshold as
\begin{equation}
    \tau(c) = \frac{\zeta_c}{\mathop{\max}_{c \in \mathcal{C}_i} \zeta_c} \cdot \eta (\mathcal{C}_i),  \ 
     c = 1, \dots, K,
\end{equation}
where the $c$-th class belongs to the set $\mathcal{C}_i$ (i.e., $c\in \mathcal{C}_s$ or $\mathcal{C}_u$). The learning condition $\eta$ distinguishes between seen and novel classes, while the class-wise condition $\zeta$ adjusts for class-wise differences. Ultimately, the confidence objective is:
\begin{equation}
\begin{split}
        \mathcal{L}_{conf} = \frac{1}{N}\sum_{i=1}^{N}& \mathbb{I}(\max(\mathbf{p}^{(i)}) > \tau(\hat{p}^{(i)}))   \cdot  H(\hat{\mathbf{p}}^{(i)}, g_\theta(\mathcal{A}(\mathbf{x}^{(i)}))).
\end{split}
    \label{eq:conf}
\end{equation}

Together with the supervised objective $\mathcal{L}_{sup} = \frac{1}{N^l} \sum_{i=1}^{N^l} H(\mathbf{y}_{gt}^{(i)}, \mathbf{p}^{(i)})$, the overall objective for OwMatch is 
\begin{equation}
  \mathcal{L} = \mathcal{L}_{sup} + \mathcal{L}_{cls} +  \mathcal{L}_{conf}.
  \label{eq:final_objective}
\end{equation}

\section{Theoretical analysis of conditional self-labeling}
\label{sec:theoretical}
To illustrate the superiority of conditional self-labeling over unconditional, we evaluate their estimators of the class distribution on unlabeled data through rigorous statistical analysis. This transformation is justified as both self-labeling methods produce corresponding self-label assignments, each representing their estimation of the class distribution on unlabeled data.

\paragraph{Formulation.}
Assuming that the class distribution of real-world data conforms to prior information $\boldsymbol{\mathcal{P}} = [p_1, p_2, \cdots, p_{K}]$.
Suppose real-world data is composed of recognized labeled data and unrecognized unlabeled data, conforming to unknown class distribution  $\boldsymbol{\mathcal{P}}^l = [p_1^l, p_2^l, \cdots, p_{K}^l]$ and $\boldsymbol{\mathcal{P}}^u = [p_1^u, p_2^u, \cdots, p_{K}^u]$ respectively. We independently sample $N=N^l+N^u$ instances from recognized and unrecognized data, respectively. Suppose $N_i=N^l_i+N^u_i$ is composed of two random variables that denote the number of recognized and unrecognized samples belonging to the $i$-th class. Obviously, we have $N^l = \sum_{i=1}^{K} N^l_i$ and $N^u = \sum_{i=1}^{K} N^u_i$.

\paragraph{Objective.}
We hope to estimate the unknown class distribution $\boldsymbol{\mathcal{P}}^u$ with $\hat{\boldsymbol{\mu}}$ based on prior information $\boldsymbol{\mathcal{P}}$ and observations of $N^l_1,N^l_2,\cdots,N^l_{K}$, then evaluate $\hat{\boldsymbol{\mu}}$ from unbiasness and ECS. Evaluation on both metrics requires estimating the number of samples in each class, denoted by $\mathbf{A} = (A_1, A_2,\dots, A_{K})$. Two self-labeling approaches (unconditional and conditional) can optimize self-label assignment, therefore obtaining two approximations of $\mathbf{A}$, denoted by $\mathbf{A}_{\text{uncon}}$ and $\mathbf{A}_{\text{con}}$. Denote the corresponding estimators as $\hat{\boldsymbol{\mu}}_{\text{uncon}}$ and $\hat{\boldsymbol{\mu}}_{\text{con}}$.

\begin{assumption}
Assume that all drawn samples with a static number of samples and class distribution follow the multinomial distribution as follows,
\begin{alignat*}{2}
    &N_1,N_2,\cdots,N_{K} &&\sim \text{Multinomial}(N,\boldsymbol{\mathcal{P}})\\
    &N^l_1,N^l_2,\cdots,N^l_{K} &&\sim \text{Multinomial}(N^l,\boldsymbol{\mathcal{P}}^l)\\
    &N^u_1,N^u_2,\cdots,N^u_{K} &&\sim \text{Multinomial}(N^u,\boldsymbol{\mathcal{P}}^u).
\end{alignat*}
Given the independency between $N^l_i$ and $N^u_j$. We basically have:
\begin{equation}
\begin{aligned}
& \mathbb{E}[N_i] = \mathbb{E}[N^l_i] + \mathbb{E}[N^u_i]\quad \forall i,j\\
& Np_i = N^lp^l_i + N^up^u_i.
\end{aligned}
\label{eq:basic}
\end{equation}
\end{assumption}

\begin{lemma}
\label{lemma:chi}
Suppose we want to test the null hypothesis ($H_0$) that categorical data $N_1, N_2,\cdots, N_{\mathcal{C}}$ come from a multinomial distribution with $K$ classes and class probability of $\boldsymbol{\mathcal{P}}$. A chi-square statistic can be constructed to test the deviation between the observations $n_1,\cdots,n_{K}$ and expected outcomes for each class.
\begin{equation}
    \chi^2 = \sum_{i=1}^{K}\frac{(n_i-\mathbb{E}_{\boldsymbol{\mathcal{P}}}[N_i]^2)}{\mathbb{E}_{\boldsymbol{\mathcal{P}}}[N_i]}\sim \chi^2_{K-1},
\end{equation}
where $\mathbb{E}_{\boldsymbol{\mathcal{P}}}[\cdot]$ denotes the population expectation of random variable. A lower chi-square value suggests that the observed data are consistent with $H_0$. Conversely, an exceedingly high chi-square value implies that either $H_0$ is incorrect or an event of low probability has happened.
\end{lemma}

Details of the above lemma are presented in Appendix~\ref{app:proof}. Then, we define the following metric to evaluate the goodness of fit of estimation based on chi-square statistics.

\begin{definition}[Expectation of chi-square statistics (ECS)]
The expectation of chi-square statistics (ECS) for $\hat{\boldsymbol{\mu}}$ are defined as the population deviation between the estimator of unlabeled class distribution $\hat{\boldsymbol{\mu}}$ and its true distribution $\boldsymbol{\mathcal{P}}^u$:
\begin{equation}
  \mathrm{ECS}(\hat{\boldsymbol{\mu}}):=\mathbb{E}[\chi^2(\mathbf{A})] =\mathbb{E}\left[ \sum_{i=1}^{K} \frac{(A_i-\mathbb{E}_{\boldsymbol{\mathcal{P}}}[N^u_i])^2}{\mathbb{E}_{\boldsymbol{\mathcal{P}}}[N^u_i]}\right],
\end{equation}
where $\mathbf{A}$ are estimators based on $N^l_1,N^l_2,\cdots,N^l_{K}$, thus are still random variables.
\end{definition}

Now, we introduce two main theorems and demonstrate the superiority of our conditional self-labeling.

\begin{theorem}
\label{thm:unbias}
Consider two estimators for class distribution on unlabeled data, $\boldsymbol{\mu}_{\mathrm{uncon}}$ and $\boldsymbol{\mu}_{\mathrm{con}}$, we have  $\boldsymbol{\mu}_{\mathrm{uncon}}$ is a biased estimator and $\boldsymbol{\mu}_{\mathrm{con}}$ is an unbiased estimator.
\end{theorem}

\begin{theorem}
\label{thm:variablity}
Suppose $r_i: = \frac{N^l\cdot p^l_i}{N}$ denote the ratio of label samples of the $i$-th class to the whole samples, $r:= \sum_i r_i$ denotes the ratio of labeled samples to the whole samples. For unlabeled sample size $N^u$, if $\sqrt{N^u}>\frac{1}{\max (|r_i-r\cdot p_i^u|,r\cdot p_j)}$ for $ \forall i\in{\mathcal{C}_l}, \forall j\in \mathcal{C}_u$, then  $\mathrm{ECS}(\hat{\boldsymbol{\mu}}_{\mathrm{con}})\leq \mathrm{ECS}(\hat{\boldsymbol{\mu}}_{\mathrm{uncon}}).$
\end{theorem}
Following rigorous statistical analysis, the generated self-label assignments from the conditional labeling method are closer to the true class distribution in the following scenarios:
\begin{itemize}
\vskip -0.2 in
    \item Estimation based on large unlabeled sample size ($N^u$);
    \item The difference between prior distribution $\boldsymbol{\mathcal{P}}$ and class distribution of unlabeled data $\boldsymbol{\mathcal{P}}^u$ is not negligible.
    \vskip -0.2 in
\end{itemize}

\section{Experiments}
\label{sec:exp}

This section presents a comprehensive evaluation of our approach. It includes experimental results and in-depth analysis, demonstrating the effectiveness of our approach.

\subsection{Experimental setup}
\label{sec:expsetup}

\textbf{Datasets.} We evaluate our approach on CIFAR-10/100~\cite{cifar}, ImageNet-100~\cite{imgnet100} and Tiny ImageNet~\cite{tinyimagenet}. A detailed description of these datasets is provided in Appendix~\ref{app:dataset}. Specifically, ImageNet-100 dataset contains 100 classes sub-sampled from ImageNet-1k following~\cite{vangansbeke2020scan}. On all datasets, we first split all classes into seen and novel classes with a \emph{novel class ratio}. Subsequent experiments will adopt a novel class ratio of 50\% unless otherwise specified. Then, we will randomly assign labels to a portion of the data from the seen classes according to the specified \emph{label ratio}, while the remaining data, along with all samples from the novel classes, are assigned to the unlabeled set.

\textbf{Implementation details.}
For a fair comparison, we apply ResNet-50~\cite{resnet} as the backbone model for ImageNet-100 and ResNet-18 for other benchmarks. We train the model with a batch size of 256 for Tiny ImageNet and 512 for other benchmarks. Following \cite{guoRobustSemiSupervisedLearning2022}, experiments across all benchmarks are implemented based on the pre-trained model from SimCLR \cite{chenSimpleFrameworkContrastive2020}. We jointly optimize backbone and prototype parameters using the standard Stochastic Gradient Descent (SGD) with momentum. We apply the cosine annealing learning rate schedule for all experiments.
Techniques including multi-crop and queue structure \cite{caronUnsupervisedLearningVisual2020} are employed to enhance the clustering objective. Additionally, RandAugment \cite{cubuk2019randaugment} serves as the strong augmentation for confidence objective. Additional implementation details are available in the Appendix~\ref{app:implement}.

\textbf{Evaluation metric.}
In assessing the efficacy of OwMatch, we adopt a multifaceted approach to evaluate accuracy following \cite{caoOpenWorldSemiSupervisedLearning2022}. Evaluation metrics include the standard accuracy for seen classes and the clustering accuracy for novel classes and all classes. Here, we leverage the Hungarian algorithm \cite{Kuhn1955TheHM} to align the predicted class assignment for novel-class instances with their ground-truth labels to obtain clustering accuracy. We also report the joint clustering accuracy across all classes using the Hungarian algorithm.

\subsection{Main results}
\begin{table}[tbp]
    \footnotesize
    \centering
    \caption{Average accuracy on the CIFAR-10/100 and ImageNet-100 with both novel class ratio and label ratio of 50\%. We compare OwMatch with existing literature on OwSSL. Also compared with other related approaches of traditional SSL, OSSL, and NCD approaches following \cite{caoOpenWorldSemiSupervisedLearning2022}. Proper modifications are made to make these approaches compatible with OwSSL; the details are in Appendix~\ref{app:baseline}. The results are averaged over three independent runs. The baseline figures are sourced from the respective papers.}
\begin{tabular}{@{}lcccccccccc@{}}
    \toprule
    \textbf{Method} & \multicolumn{3}{c}{\textbf{CIFAR-10}} & \multicolumn{3}{c}{\textbf{CIFAR-100}} & \multicolumn{3}{c}{\textbf{ImageNet-100}}\\
    \cmidrule(lr){2-4} \cmidrule(l){5-7}  \cmidrule(lr){8-10} 
        & Seen  & Novel & All   & Seen  & Novel & All & Seen  & Novel & All \\
    \midrule
    FixMatch \cite{sohnFixMatchSimplifyingSemiSupervised2020}  & 71.5  & 50.4  & 49.5  & 39.6  & 23.5  & 20.3  & 65.8  & 36.7  & 34.9  \\
    DS\textsuperscript{3}L \cite{guoSafeDeepSemiSupervised2020} & 77.6  & 45.3  & 40.2  & 55.1  & 23.7  & 24.0  & 71.2  & 32.5  & 30.8  \\
    CGDL \cite{Sun_2020_CVPR}  & 72.3  & 44.6  & 39.7  & 49.3  & 22.5  & 23.5 & 67.3 & 33.8  & 31.9  \\
    DTC \cite{han2019learning} & 53.9  & 39.5  & 38.3  & 31.3  & 22.9  & 18.3  & 25.6 & 20.8  & 21.3  \\
    RankStats \cite{han2020automatically} & 86.6  & 81.0  & 82.9  & 36.4  & 28.4  & 23.1  & 47.3  & 28.7  & 40.3  \\
    SimCLR \cite{chenSimpleFrameworkContrastive2020}  & 58.3  & 63.4  & 51.7  & 28.6  & 21.1  & 22.3  & 39.5 & 35.7  & 36.9  \\
    UNO \cite{fini2021unified} & 91.6  & 69.3  & 80.5  & 68.3  & 36.5  & 51.5 & -  & -  & -   \\
    ORCA \cite{caoOpenWorldSemiSupervisedLearning2022} & 88.2  & 90.4  & 89.7  & 66.9  & 43.0  & 48.1  & 89.1  & 72.1  & 77.8  \\
    NACH \cite{guoRobustSemiSupervisedLearning2022} & 89.5  & 92.2  & 91.3  & 68.7 & 47.0  & 52.1  & 91.0  & 75.5  & 79.6   \\
    OpenLDN \cite{rizveOpenLDNLearningDiscover2022} & 95.7  & 95.1  & 95.4  & 73.5  & 46.8  & 60.1  & 89.6  & 68.6  & 79.1  \\
    TRSSL \cite{rizve2022realistic} &  \textbf{96.8}  & 92.8  & 94.8  & 80.0  & 49.3  & 64.7 & -  & -  & -   \\
    OpenCon \cite{sun2023opencon} & 89.3  & 91.1  & 90.4  & 69.1  & 47.8  & 52.7   & 90.6 & 80.8  & 83.8 \\
    \midrule
    OwMatch &  93.0  & 95.9  & 94.4  & 74.5  & 55.9  & 65.1  & \textbf{91.7}  & 72.0  & 81.8 \\
    OwMatch+ & 96.5  & \textbf{97.1}  & \textbf{96.8}  & \textbf{80.1}  & \textbf{63.9}  & \textbf{71.9}  & 91.5  & 79.6  & \textbf{85.5} \\
    \bottomrule
    \end{tabular}%
    \vskip -0.1 in
  \label{tab:main50}%
\end{table}%

We consider and evaluate two versions of our method, called OwMatch and OwMatch+. OwMatch represents the standard version as illustrated in Figure~\ref{fig:pipline}, while OwMatch+ incorporates the multi-crop technique for additional augmentation. Detailed distinctions between the two versions are provided in the Appendix \ref{app:implement}.
We evaluate our method on all benchmarks using a label ratio of 10\% and 50\% with the comprehensive experiment results provided in Table~\ref{tab:main50}, \ref{tab:main10}, and \ref{tab:tinyimage}. Results in Table~\ref{tab:main50} show that OwSSL approaches significantly outperform current state-of-the-art methods in traditional SSL, OSSL, and NCD by a considerable margin.
On the other hand, OwMatch achieves state-of-the-art across all benchmarks and evaluation metrics. It can not only classify novel classes accurately but also maintain robust performance on seen classes. On CIFAR-10, we observed OwMatch outperforms OpenLDN on novel and all classes by 2.0\% and 1.4\%, respectively. It is noteworthy that the enhancement brought about by OwMatch is more pronounced on the CIFAR-100 dataset, which presents a greater challenge due to the increasing number of classes. Regarding CIFAR-100, our method surpasses TRSSL by approximately 14.6\% on novel classes and 7.2\% on all classes. Subsequently, we extend to evaluate ImageNet-100 and observe a similar trend, with OwMatch+ showing significant improvement of \textbf{1.7\%} on all-class accuracy compared to previous state-of-the-art approaches.

\paragraph{Principle analysis of conditional self-labeling.} OwMatch primarily relies on high-quality self-label assignment to alleviate the model's confirmation bias. To clearly illuminate this progress during training, we employ the Manhattan distance $\sum_{i\in K} |c_i-c^{gt}_i|$ as a metric to evaluate the bias between the considered class distribution $\{c_i\}_{i=1}^K$ and the ground truth $\{c^{gt}_i\}_{i=1}^K$. Table \ref{tab_rebutt:disc1} demonstrates the debiasing process: the model's confirmation bias is pronounced in the early epochs, whereas the bias of optimized self-label assignment is relatively minor. As training advances, the self-label assignment continues to guide the model, effectively mitigating the confirmation bias, as reflected in the decreasing $B_m$ and the absolute difference between $B_m$ and $B_s$.

\vspace{-1em}
\begin{table}[h]
    \footnotesize
    \centering
    \caption{The Manhattan distance (MD) is used to evaluate the confirmation bias. The first row presents the bias between the model's predictive class distribution and the ground truth, denoted as $B_m$. The second row reflects the bias between the self-label assignment and the ground truth, denoted as $B_s$. The third row computes the absolute difference between $B_m$ and $B_s$, highlighting the debiasing effect of high-quality self-label assignments.}
\begin{tabular}{@{}lccccccccccc@{}}
    \toprule
  \textbf{Bias of} & \textbf{Epoch 1} & \textbf{Epoch 2} & \textbf{Epoch 3} &\textbf{Epoch 6} & \textbf{Epoch 10} & \textbf{Epoch 30} &\textbf{Epoch 50}\\
    \midrule
      $B_m$ & 0.4463 & 0.2939 & 0.2474 & 0.1753 & 0.1505 & 0.0904 & 0.0798  \\
      $B_s$ &  0.1004 & 0.0754 & 0.0893 & 0.0613 & 0.0407 & 0.0255 & 0.0219  \\
      $|B_m - B_s|$ & 0.3459 & 0.2185 & 0.1581 & 0.1140 & 0.1098 & 0.0649 & 0.0579 \\ 
    \bottomrule
    \end{tabular}%
  \label{tab_rebutt:disc1}%
  \vspace{-1em}
\end{table}%

\subsection{Ablations, analysis, and real-scenario applications}
\label{sec:ablation}

To investigate the impact of each component, we embark on comprehensive ablation studies with both novel class ratio and label ratio of 50\%. The first row in Table~\ref{tab:ablation} showcases the foundational model performance, whose objective consists of only \emph{unconditional} clustering objective and supervised objective, already achieving impressive performance. We then analyze the effect of integrating a conditional self-labeling framework on CIFAR-100, which boosts novel-class accuracy by 1.0\% on average. Additionally, the positive impact of consistency regularization is observed: roughly 0.9\% enhancement across all evaluation metrics. Our ablation studies highlight the essential contribution of each component in OwMatch. Individually, each plays a significant part in the intended functionality, and together, these elements coalesce into a cohesive and robust framework. We also ablate other factors, including the number of local views for clustering objectives and iterations for the Sinkhorn-Knopp algorithm, with detailed statements provided in Table~\ref{tab:crops} and \ref{tab:iteration}.

\vspace{-1em}
\begin{table}[htbp]
  \footnotesize
  \centering
  \caption{Ablation studies on each component with both novel class ratio and label ratio of 50\%. Here, \textbf{ConSL} refers to conditional self-labeling, \textbf{PLCR} refers to pseudo-label consistency regularization, and \textbf{OwHT} refers to an open-world hierarchical thresholding scheme.}
  \begin{tabular}{cccccccccccc}
    \toprule
    \multicolumn{3}{c}{\textbf{Components}} & \multicolumn{3}{c}{\textbf{CIFAR-10}} & \multicolumn{3}{c}{\textbf{CIFAR-100}} & \multicolumn{3}{c}{\textbf{Tiny-ImageNet}} \\
    \cmidrule(lr){1-3} \cmidrule(l){4-6} \cmidrule(l){7-9} \cmidrule(l){10-12}
    ConSL  & PLCR & OwHT & Seen  & Novel & All & Seen  & Novel & All & Seen  & Novel & All \\
    \midrule
    \texttimes  & \texttimes  & \texttimes  & 96.5  & 90.2  & 93.3 & 78.8  & 56.7  & 67.7 & 66.5 & 38.1 & 52.0 \\
    \checkmark  & \texttimes  & \texttimes  & 95.4  & 96.4  & 95.9 & 79.2  & 58.5  & 68.7 & 66.0 & 39.4 & 52.4 \\
    \checkmark  & \checkmark  & \texttimes  & 96.3  & 97.3  & 96.8 & 80.1  & 59.4  & 69.6 & 68.6 & 42.0 & 54.2 \\
    \texttimes  & \checkmark  & \checkmark  & 97.1  & 90.4  & 93.8 & 80.7  & 59.7  & 69.9 & 69.7 & 41.4 & 54.6 \\
    \checkmark  & \checkmark  & \checkmark  & 96.5  & 97.1  & 96.8 & 80.1  & 63.9  & 71.9 & 68.8 & 42.4 & 55.0 \\
    \bottomrule
    \end{tabular}%
  \label{tab:ablation}%
\end{table}%

\paragraph{Comparison study on varying thresholding strategies.} We compare our proposed open-world hierarchical thresholding approach with two prominent techniques: static thresholding \cite{sohnFixMatchSimplifyingSemiSupervised2020} and self-adaptive 
 thresholding \cite{wangFreeMatchSelfadaptiveThresholding2023}. As illustrated in Table \ref{tab_rebuttal:freematch_openfree}, our proposal achieves superior performance in both novel- and all-class clustering accuracy across varying thresholding techniques. While self-adaptive has proven effective under closed-world scenarios, it encounters challenges in open-world settings. The pronounced disparity in overall learning conditions between seen and novel classes, as illustrated in Figure \ref{fig:thres}, can lead to unstable global thresholds. The class-wise adaptive approach based on that may exaggerate this issue, resulting in suboptimal performance. We implement a hierarchical structure to mitigate the instability sourcing from distinct learning dynamics of seen and novel classes. 

\vspace{-1em}
\begin{table}[h]
\begin{minipage}{.49\textwidth}
	\caption{Performance comparison of static, self-adaptive, and our OwHT thresholding techniques on CIFAR-100 with both novel class ratio and label ratio of 50\%.}
        \label{tab_rebuttal:freematch_openfree}%
	\fontsize{9}{9}\selectfont
	\setlength\tabcolsep{3pt} 
	\centering
    \begin{tabular}{@{}lcccc@{}}
    \toprule
    \textbf{Thresholding} & Seen Acc & Novel Acc & All Acc \\
    \midrule
    Static - 0.7 & 80.1 & 59.4 & 69.6 \\
    Static - 0.8 & 79.8 & 63.9 & 71.7 \\
    Static - 0.9 & 80.2 & 62.8 & 71.3 \\
    \midrule
    Self-adaptive & {\bf 81.0} & 60.5 & 70.6 \\
    OwHT & 80.1 & {\bf 63.9} & {\bf 71.9} \\
    \bottomrule
    \end{tabular}%
	\end{minipage}
	\hfill
	\begin{minipage}{.49\textwidth}
	\caption{Comparison with GCD-related works: average accuracy on the ImageNet-100 with both novel class ratio and label ratio of 50\%.}
	\label{tab_rebutt:main}
	\fontsize{9}{9}\selectfont
	\setlength\tabcolsep{3pt} 
	\centering
    \begin{tabular}{@{}lcccccccccc@{}}
    \toprule
    \textbf{Method} & \textbf{Backbone} & Seen  & Novel & All \\
    \midrule
    GCD \cite{gcd2022} & ViT-B/16 & 91.8 & 63.8 & 72.7  \\
    SimGCD \cite{simgcd} & ViT-B/16 &  93.1 & 77.9 & 83.9  \\
    InfoSieve \cite{InfoSieve} & ViT-B/16 & 84.9 & 78.3 & 80.5  \\
    CiPR \cite{cipr} & ViT-B/16 & 84.9 & 78.3 & 80.5 \\
    PromptCAL \cite{promptcal} & ViT-B/16 &  92.7 & 78.3 & 83.1 \\
    \midrule
    OwMatch+ & ResNet-50 &  91.5  & \textbf{79.6} & \textbf{85.5} \\
    \bottomrule
    \end{tabular}%
	\end{minipage}
	\vspace{-0.5em}
\end{table}

\paragraph{On the comparison between OwSSL and Generalized Category Discovery.} The OwSSL setting resembles the subsequently proposed Generalized Category Discovery (GCD) setting \cite{gcd2022}, with both assuming the existence of novel classes and that a portion of the data is labeled for seen classes. However, there are notable differences between these two groups of methods: 1) GCD-related methods leverage supervised contrastive learning \cite{supcon} on labeled data and self-supervised contrastive learning \cite{chenSimpleFrameworkContrastive2020} on all training data, whereas OwSSL typically employs pairwise similarity-based methods for clustering samples; 2) GCD-related works typically employ a pre-trained ViT-Base/16 backbone, which has significantly more parameters than the ResNet-18 or ResNet-50 models commonly used in OwSSL methods.

It is unfair to compare these two types of methods directly. Here, we still include a comparison with those GCD-related works to demonstrate the effectiveness of our method. Table \ref{tab_rebutt:main} shows that our method outperforms existing approaches in novel-class and all-class accuracy on ImageNet-100 despite using a simpler model.

\paragraph{Ablation study on supervision components in the overall objective.} The overall objective of OwMatch consists of a standard supervised objective, clustering objective, and confidence objective. Both the supervised and clustering objectives involve the use of labeled data, raising concerns about overlap in functionality. To investigate the significance of each component, we conduct an ablation study in which we modify the overall objective in two ways: 1) removing the supervised objective and 2) excluding labeled data from the online clustering process. The results are reported in Table \ref{tab_rebuttal:comparison_super_labelcluster}.

In the first case, we observe a decrease in seen-class accuracy while maintaining novel-class clustering performance, while the latter case exhibits the opposite tendency: seen-class accuracy remains high, but novel-class clustering accuracy declines. In comparison to the previous cases, our overall objective integrates both components to strike a balance between clustering and confidence. The supervised objective enhances seen-class accuracy through one-hot supervision, while the clustering objective with conditional self-labeling improves novel-class clustering accuracy by incorporating labeled data. This harmonious approach yields the best all-class accuracy while roughly maintaining both seen- and novel-class performance.

\vspace{-1em}
\begin{table}[h]
\begin{minipage}{.49\textwidth}
    \caption{Model performance under varying applications of supervision in the overall objective.}
    \label{tab_rebuttal:comparison_super_labelcluster}%
    \fontsize{9}{9}\selectfont
    \setlength\tabcolsep{3pt} 
    \centering
    \begin{tabular}{@{}lcccc@{}}
        \toprule
        \textbf{Objective} & Seen  & Novel & All\\
        \midrule
        w/o supervised objective & 76.8 &  64.4 & 70.6 \\
        w/o supervision for clustering&  80.3 & 61.2 & 70.7 \\
        overall objective & 80.1 & 63.9 &  71.9 \\
        \bottomrule
    \end{tabular}%
\end{minipage}
\hfill
\begin{minipage}{.49\textwidth}
    \caption{Estimation of the number of classes across benchmarks.}
    \label{tab_rebuttal:class_estimation}%
    \fontsize{9}{9}\selectfont
    \setlength\tabcolsep{3pt} 
    \centering
    \begin{tabular}{@{}lcccc@{}}
        \toprule
         & CIFAR10 & CIFAR100 & ImageNet100 \\
        \midrule
        Ground Truth & 10 & 100 & 100 \\
        Estimation & 10 & 104 & 111 \\
        Error & 0\% & 4\% & 11\% \\
        \bottomrule
    \end{tabular}%
\end{minipage}
\vspace{-0.5em}
\end{table}

The above evaluations are typically conducted under relatively ideal conditions: the datasets are class-balanced, and both the prior class distribution and the number of novel classes are available. However, in real-world scenarios, it is crucial to address the dependency on these assumptions. We will elaborate on each of these aspects to demonstrate the practical effectiveness of OwMatch.

\paragraph{Estimating the number of novel classes.} OwMatch and other baselines typically assume that the number of novel classes is pre-determined for clarity in evaluation. However, this prior knowledge is often unavailable in practice, necessitating a precise estimation of the number of novel classes in advance. We primarily follow the approaches of GCD \cite{gcd2022} and TRSSL \cite{rizve2022realistic} to estimate the number of classes. Specifically, $K$-means clustering is performed on representations of the entire dataset from the pre-trained ViT-B/16 backbone. The optimal value of $k$ is determined by evaluating the clustering accuracy on the labeled samples calculated by the Hungarian algorithm. This accuracy serves as a scoring function, optimized using Brent's algorithm to find the that maximizes performance on the labeled data. The estimation results across generic benchmarks are shown in Table \ref{tab_rebuttal:class_estimation}, which illustrates that the estimated class numbers come close to the ground truth. We also evaluate OwMatch's sensitivity to varying extents of class number estimation error, with results reported in Appendix \ref{app:results}.

\vspace{-1em}
\begin{table}[htbp]
    \footnotesize
    \centering
    \caption{Performance on generic recognition benchmarks with varying imbalance factors (IF), with and without prior class distribution. These benchmarks come with both novel class ratio and label ratio of 50\%.}
 \begin{tabular}{@{}lccccccccccc@{}}
    \toprule
    \textbf{Dataset} & \textbf{Prior} & \multicolumn{3}{c}{\textbf{Uniform (IF=1)}} & \multicolumn{3}{c}{\textbf{IF=10}} & \multicolumn{3}{c}{\textbf{IF=20}}\\
    \cmidrule(lr){3-5} \cmidrule(l){6-8}  \cmidrule(lr){9-11} 
          &  & Seen  & Novel & All   & Seen  & Novel & All & Seen  & Novel & All \\
    \midrule
    \multirow{2}{7em}{\textbf{CIFAR-10}} & w/ &  96.5  & 97.1  & 96.8  & 93.7  & 72.1  & 82.5  & 92.9  & 70.1  & 80.9 \\
     & w/o & 96.9  & 90.9  & 93.9  & 95.8  & 66.5  & 80.3  & 95.3  & 64.2  & 78.8 \\
    \multirow{2}{7em}{\textbf{CIFAR-100}} & w/ & 80.1  & 63.9  & 71.9  & 76.8  & 42.0  & 57.3  & 76.1  & 35.2  & 51.9 \\
     & w/o & 82.5  & 57.9  & 69.2  & 74.6  & 39.7  & 54.1  & 73.9  & 33.9  & 49.2 \\
    \multirow{2}{7em}{\textbf{Tiny-ImageNet}} & w/ & 68.8  & 42.4  & 55.0  & 61.7  & 25.1  & 41.6  & 62.4  & 21.7  & 38.3 \\
     & w/o & 69.6  & 40.6  & 54.8  & 61.0  & 24.9  & 40.1  & 61.3  & 20.3  & 36.9 \\
    \bottomrule
    \end{tabular}%
    \vskip -0.1 in
  \label{tab:abla_imb}%
\end{table}%

\paragraph{Data imbalance.}
Most generic benchmarks feature class-balanced, whereas real-world data tend to exhibit long-tailed class distribution. Our approach accommodate to arbitrary class distribution by constraining the optimized self-label assignment to comply with the prior class distribution, thereby naturally mitigating performance degradation caused by data imbalance. We evaluate our approach on imbalanced benchmarks, constructed with varying imbalance factors following TRSSL \cite{rizve2022realistic}. Results in Table~\ref{tab:abla_imb} demonstrate that OwMatch effectively addresses the challenge of data imbalance.

\paragraph{Training without prior.}
In scenarios where prior class distribution is unavailable, we propose an adaptive estimation scheme to make OwMatch still function \emph{without relying on any prior assumptions}. Specifically, we initially adopt class-balanced prior if no prior information is available; then, the class distribution for conditional self-labeling is estimated and continuously updated based on model prediction. Next, standard training with estimated class distribution and distribution estimation are alternately conducted, with results reported in Table~\ref{tab:abla_imb}. We observe that the reduction in all-class accuracy achieved through the adaptive estimation scheme remains within 3\% across almost all benchmarks and imbalance factors. These results reveal that the straightforward estimation technique performs robustly in the absence of prior knowledge.

\section{Conclusion}
This work integrates techniques from self-SL and SSL, refining them to present a new perspective on solving open-world SSL. We demonstrate that conditional self-labeling can achieve an unbiased estimation of the class distribution on unlabeled data with prior information, leading to high-quality self-label assignment with reduced confirmation bias. Our future endeavors will be directed toward developing solutions that are more aligned with realistic scenarios where such prior information might not be readily available or hard to be estimated. This will involve exploring methodologies that can effectively handle uncertainty and variability inherent in real-world data distributions.

\newpage

\section*{Acknowledgements}
Chao Wang would like to acknowledge the support from the National Natural Science Foundation of China under Grant 12201286, the Shenzhen Science and Technology Program 20231115165836001, Guangdong Basic and Applied Research Foundation 2024A1515012347. This research was conducted using the computing resources provided by the Research Center for the Mathematical Foundations of Generative AI in the Department of Applied Mathematics at The Hong Kong Polytechnic University.

\nocite{langley00}

\bibliography{main}
\bibliographystyle{plain}

\newpage
\appendix
\section*{Technical Appendices}

\paragraph{Roadmap of technical appendices.}
These appendices are structured as follows: Appendix~\ref{app:dataset} introduces dataset details utilized in our experiments. Appendix~\ref{app:implement} outlines the implementation specifics, including data augmentations and hyperparameters. Modification details for utilized baselines are provided in Appendix~\ref{app:baseline}. Additional experiment results consisting of supplementary main results, in-depth analysis, and ablation study of hyperparameters are reported in Appendix~\ref{app:results}. Complete and rigorous proof of theoretical results is represented in Appendix~\ref{app:proof}. Appendix~\ref{app:impact} discusses the social impacts of our work, and the limitations are considered in Appendix~\ref{app:limit}

\section{Datasets}
\label{app:dataset}

The details of the datasets utilized in our experiments are provided in Table~\ref{tab:dataset}, which includes dataset statements, as well as the corresponding backbones and batch sizes for training. The choice of backbone and batch size matches previous works \cite{caoOpenWorldSemiSupervisedLearning2022, guoRobustSemiSupervisedLearning2022} for fair comparison. For CIFAR-10/100 datasets, we employ a simple pre-processing encompassing random crop with padding and horizontal flip. To make ResNet-18 compatible with CIFAR input data with a small resolution of $32 \times 32$, we refine the CNN by setting the kernel size to $3 \times 3$ and applying a stride of 1. We train the model with a batch size of 512 over 300 epochs. For ImageNet-100 and Tiny-ImageNet datasets, raw images are pre-processed with random resized crop and horizontal flip \cite{resnet}. We use the standard version of ResNet-50 and ResNet-18 as the backbone, respectively. We leverage the standard SGD method with momentum and weight decay to optimize the network parameters; see hyperparameters in Table~\ref{tab:hyper}.

\begin{table}[htbp]
  \centering
  \caption{Details of evaluation benchmarks, we show the number of classes, dataset statistics, selected backbone, and batch size for training.}
    \begin{tabular}{@{}lcccccc@{}}
    \toprule
    \textbf{Dataset} & \textbf{Num. Class} & \textbf{Train Samples} & \textbf{Test Samples} & \textbf{Backbone} & \textbf{Batch size}\\
    \midrule
    CIFAR-10 \cite{cifar}  & 10  & 50,000  & 10,000 & ResNet-18  & 512  \\
    CIFAR-100 \cite{cifar} & 100  & 50,000  & 10,000 & ResNet-18  & 512 \\
    ImageNet-100 \cite{imgnet100} & 100  & 128,545  & 5,000 & ResNet-50  & 512 \\
    Tiny-ImageNet \cite{tinyimagenet} & 200  & 100,000 & 10,000 & ResNet-18  & 256 \\
    \bottomrule
    \end{tabular}%
  \label{tab:dataset}%
\end{table}%

\section{Implementation details}
\label{app:implement}

\paragraph{Computational resources.}
The foundational algorithm of our study is constructed utilizing Python 3.8 and PyTorch 2.1 \cite{pytorch}. All experiments are carried out on NVIDIA's Tesla A100 GPU with 40G memory. All benchmarks are public and can be easily downloaded.

\paragraph{Strong augmentation.}
Furthermore, we apply the strong augmentation to input data for all experiments following FixMatch \cite{sohnFixMatchSimplifyingSemiSupervised2020}, including random resized crop, horizontal flip, and RandAugment \cite{cubuk2019randaugment}. It should be noted that the only additional enhancement in strong augmentation is RandAugment compared to basic pre-processing. Specifically, RandAugment randomly selects transformations from a collection of options for each sample in a mini-batch. We employed the same sets of image transformations as those used in RandAugment. A complete list of these transformations can be found in the original work \cite{cubuk2019randaugment}.

\paragraph{OwMatch v.s. OwMatch+.}
We evaluate two versions of our approach in the main results, as illustrated in Table~\ref{tab:main50}, \ref{tab:main10} and \ref{tab:tinyimage}. In short, OwMatch+ further incorporates the multi-crop technique as an additional augmentation to boost clustering capacity and, hence, improve model performance. The multi-crop strategy was proposed by SwAV \cite{caronUnsupervisedLearningVisual2020} to additionally augment images by covering only small sections. The resulting low-resolution images, referred to as local views, allow more augmentations at a marginally increased computational cost. Compared to simple preprocessing (global views), local views generated through a multi-crop strategy involve resized cropping at smaller scales but in greater numbers, as illustrated in Table~\ref{tab:mc}. And they experience additional color distortion \cite{chenSimpleFrameworkContrastive2020} consisting of random color jitters, solarizing, and equalization in pursuit of model robustness. Here we apply the multi-crop technique to produce many low-resolution images (local views) and set the hyperparameters following PAWS~\cite{assranSemiSupervisedLearningVisual2021}, as detailed in Table~\ref{tab:mc}.

\begin{table}[htbp]
  \centering
  \caption{The details of crop augmentation hyperparameters.}
    \begin{tabular}{@{}lcccccc@{}}
    \toprule
    \textbf{Dataset} & \multicolumn{2}{c}{\textbf{Global views}} & \multicolumn{4}{c}{\textbf{Local views}} \\
    \cmidrule(lr){2-3} \cmidrule(l){4-7}
          & Crop scale  & Resize   & Crop scale  & Resize & Numbers & Distortion \\
    \midrule
    CIFAR-10  & (0.75,1)  & 32   & (0.3,0.75)  & 18  & 4 & 0.5 \\
    CIFAR-100 & (0.75,1)  & 32  & (0.3,0.75)  & 18  & 4 & 0.5 \\
    ImageNet-100 & (0.2,1)  & 224  & (0.14,0.3)  & 18  & 4 & 1\\
    Tiny-ImageNet & (0.75,1)  & 64  & (0.3,0.75)  & 36  & 4 & 1\\
    \bottomrule
    \end{tabular}%
  \label{tab:mc}%
\end{table}%

Except for additional multi-crop augmentations, OwMatch and OwMatch+ differ slightly in the form of clustering objective. Specifically, we augment the input image by taking 2 full-resolution crops (normally and strongly augmented global views) and $V$ low-resolution crops (local views). Note that we optimize the self-label $\tilde{\mathbf{q}}$ with only global views, since local views can only capture localized semantic information and are unable to provide a comprehensive overview of the entire image. We promote the model consistency by encouraging the prediction of different local views to be close to the optimized self-labels. Specifically, the clustering objective of OwMatch+ is formulated by
\begin{equation}
    \label{eq:cls_owmatch+}
  \mathcal{L}_{cls}^+ = \frac{1}{(V+1) N} \sum_{i=1}^N \left[\sum_{v=1}^V H(\tilde{\mathbf{q}}^{(i)}, g_{\theta}(\alpha_v(\mathbf{x}^{(i)})))+ H(\tilde{\mathbf{q}}^{(i)}, \mathbf{p}^{(i)})\right],
\end{equation}
where $\tilde{\mathbf{q}}^{(i)}$ correspond to the optimized self-label of global view, and $g_{\theta}(\alpha_1(\mathbf{x}^{(i)})),\cdots, g_{\theta}(\alpha_V(\mathbf{x}^{(i)}))$ stand for predictions of $V$ local views. Increasing the number of random low-resolution crops encourages the model to learn global-to-local information \cite{assranSemiSupervisedLearningVisual2021}, which reflects in performance gain across all benchmarks; see main results in Table~\ref{tab:main50}, \ref{tab:main10} and \ref{tab:tinyimage}. The utilization of low-resolution images boosts the model's efficiency with only a marginal rise in computational costs.

The effectiveness of conditional self-labeling is significantly compromised when the ratio of batch size $(B)$ to class numbers $(K)$ is relatively small. 
In a scenario where a class is disproportionately sampled in the labeled data of a batch, the conditional self-labeling mechanism might be unable to reassign the unlabeled data to that particular class. Notably, when this ratio falls below 1, assigning $B$ samples to all $K$ classes becomes unfeasible. Therefore, we leverage the queue structure to store data from previous batches by following SwAV \cite{caronUnsupervisedLearningVisual2020}. In practice, a queue of 1024 logits are stored for the implementation of the Sinkhorn-Knopp algorithm, which is utilized to derive the self-label assignment. We then retain the logits from the last batch of the optimized self-labels to construct the clustering objective. Such queue length proves effective in our experiments with a large batch size (e.g., 512) and a relatively small number of categories (e.g., 100 classes for CIFAR-100). However, when dealing with high-resolution images that encompass a greater variety of categories, storing much more previous data information is essential.

\paragraph{Hyperparameters}
Here, we provide a comprehensive list of hyperparameters in Table~\ref{tab:hyper}. For hyperparameters related to the SGD optimizer, we adhere to the settings used in the previous works \cite{caoOpenWorldSemiSupervisedLearning2022, rizve2022realistic} to ensure a fair comparison. Regarding hyperparameters introduced by our proposed method, we perform ablation studies to determine the most appropriate values, specifically for SK-iteration and the number of local views, as detailed in Appendix~\ref{app:results}. These hyperparameters are selected based on a balanced consideration of computational costs and model performance.

\begin{table}[htbp]
  \centering
  \caption{List of hyperparameters for CIFAR-10/100, ImageNet-100, and Tiny-ImageNet.}
    \begin{tabular}{@{}l|cccc@{}}
    \toprule
    \textbf{Hyper-parameter} & CIFAR-10 & CIFAR-100 & ImageNet-100 & Tiny-ImageNet \\
    \midrule
    SGD-momentum  & \multicolumn{4}{c}{0.9}  \\
    SGD-learning rate  & \multicolumn{4}{c}{0.1}  \\
    SK-$\epsilon$ & \multicolumn{4}{c}{10}  \\
    SK-iteration & \multicolumn{4}{c}{10}  \\
    \# of local views & \multicolumn{4}{c}{4}  \\
    \midrule
    SGD-weight decay & \multicolumn{1}{c|}{0.0005}  & \multicolumn{1}{c|}{0.0005}  & \multicolumn{1}{c|}{0.0001} & 0.0001 \\
    \bottomrule
    \end{tabular}%
  \label{tab:hyper}%
\end{table}%

\section{Baseline implementation details}
\label{app:baseline}

We compare our proposal with baselines from other settings: traditional SSL, open-set semi-supervised learning (OSSL), and novel class discovery (NCD). We will elaborate on the modifications to these settings separately. Traditional SSL methods cannot deal with the novel-class instances and we extend it in the following manner: samples are firstly divided into seen-class and novel-class instances based on out-of-distribution (OOD) criteria, then we report the standard classification accuracy on seen-class instances and apply K-means algorithms to achieve clustering accuracy on the novel-class instances. Hungarian algorithm \cite{Kuhn1955TheHM} is utilized to match the clustering result and their ground-truth labels, this result is reported as novel-class accuracy. For the traditional SSL method, FixMatch \cite{sohnFixMatchSimplifyingSemiSupervised2020}, we separate the OOD samples based on confidence scores produced by the Softmax function. Many OSSL approaches like CGDL \cite{Sun_2020_CVPR} are naturally capable of detecting outliers, thus we directly cluster the considered outliers by the K-means algorithm and report the clustering accuracy without manually inspecting OOD samples. Since DS3L \cite{guoSafeDeepSemiSupervised2020} applies the re-weighting technique to downsize the passive effect of OOD samples, we consider the samples with the lowest weight as outliers. Both traditional SSL and weighting-based OSSL approaches depend on the OOD likelihood score to partition the inliers and outliers. Here we follow ORCA \cite{caoOpenWorldSemiSupervisedLearning2022} to determine the threshold for OOD samples by using ground-truth class information.

NCD methods are trained to discover novel classes in unlabeled data with totally novel-class instances, thus failing to recognize seen-class instances. For NCD approaches without seen-class classification heads, like DTC \cite{han2019learning} and RankStats \cite{han2020automatically}, we report the performance on novel classes and extend them to classify seen classes by assuming the seen-class instances in unlabeled data as novel. Then we extend the unlabeled classification head to include logits for seen classes by following \cite{caoOpenWorldSemiSupervisedLearning2022} and leverage the Hungarian algorithm to match the discovered classes with ground-truth labels within labeled data, the best assignment is reported as seen-class accuracy. And for recent UNO~\cite{fini2021unified} with explicitly labeled classification heads, we generate pseudo-labels for both seen- and novel-class instances based on model predictions from concatenated labeled and unlabeled classification heads. Therefore, both seen and novel class classification accuracy can be directly computed and reported. We apply the same pre-trained model on NCD methods to demonstrate that the enhanced performance is not attributable to the the application of pre-training. Additionally, we present the clustering outcomes based on representations from the pre-trained model.

\section{Additional results}
\label{app:results}
\begin{figure}[htbp]
\centering
\begin{subfigure}[b]{0.28\textwidth}
    \includegraphics[width=\textwidth]{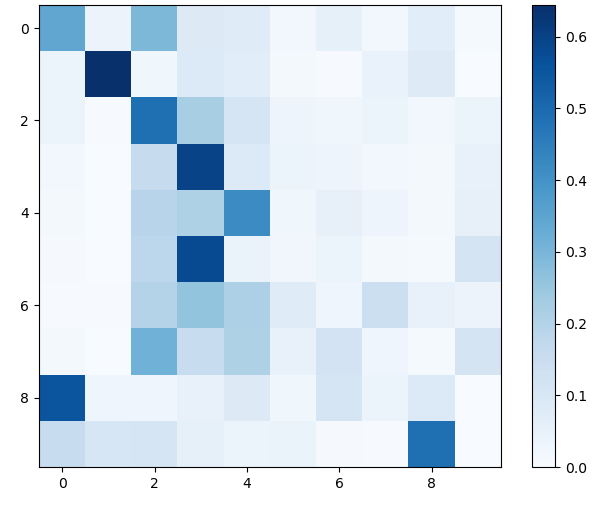}
    \caption{Epoch 1}
    \label{fig:left}
\end{subfigure}\qquad
\begin{subfigure}[b]{0.28\textwidth}
    \includegraphics[width=\textwidth]{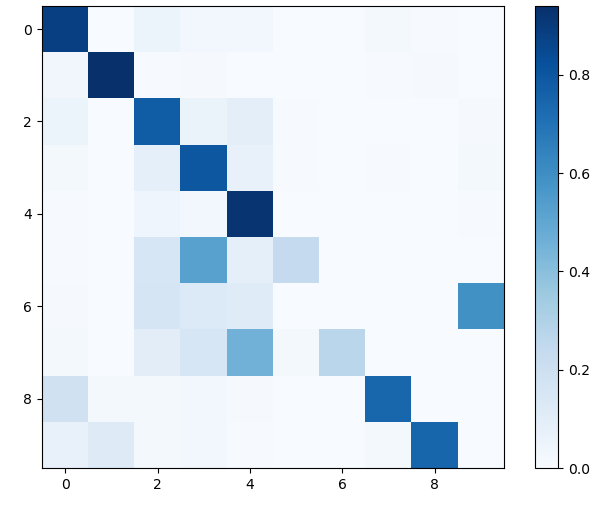}
    \caption{Epoch 10}
    \label{fig:middle}
\end{subfigure}\qquad
\begin{subfigure}[b]{0.28\textwidth}
    \includegraphics[width=\textwidth]{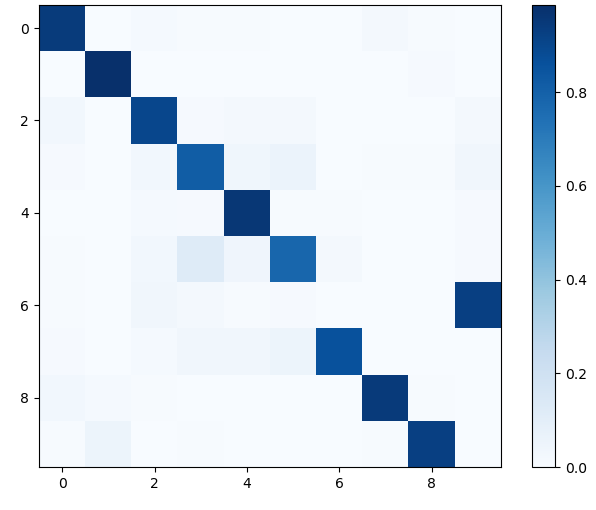}
    \caption{Epoch 100}
    \label{fig:right}
\end{subfigure}
\caption{Confusion matrices on CIFAR-10 with both novel class ratio and label ratio of 50\%. The model needs to classify the initial five seen classes accurately (as reflected in the diagonal elements). While for the novel classes (6-10), the classes clustering are required to align with the ground-truth label (dark blue in one cell).}
\label{fig:confusion}
\vspace{-1em}
\end{figure}

\paragraph{Training process.}
We plot the confusion matrices on CIFAR-10 with both novel class ratio and label ratio of 50\% in Figure~\ref{fig:confusion}. This collection of images compellingly demonstrates that the bias derived from novel classes is progressively mitigated as the experiment advances, leading to continuous improvement in the model's prediction accuracy.

As depicted in Figure~\ref{fig:left}, at the beginning of training, the model struggles to effectively distinguish between novel and seen class instances, although it can classify seen class instances normally. As training progresses, an increasing number of samples are assigned to the novel classes and prediction accuracy for the seen-class instances advances, as illustrated in Figure~\ref{fig:middle}. In the later stages of training as shown in Figure~\ref{fig:right}, the model becomes capable of accurately classifying the seen-class instances and clustering novel-class instances simultaneously.

\begin{table}[htbp]
    \footnotesize
    \centering
    \caption{Average accuracy on the CIFAR-10/100 and ImageNet-100 with novel class ratio of 50\% and labeled ratio of 10\%.}
 \begin{tabular}{@{}lcccccccccc@{}}
    \toprule
    \textbf{Method} & \multicolumn{3}{c}{\textbf{CIFAR-10}} & \multicolumn{3}{c}{\textbf{CIFAR-100}} & \multicolumn{3}{c}{\textbf{ImageNet-100}}\\
    \cmidrule(lr){2-4} \cmidrule(l){5-7}  \cmidrule(lr){8-10} 
          & Seen  & Novel & All   & Seen  & Novel & All & Seen  & Novel & All \\
    \midrule
    FixMatch \cite{sohnFixMatchSimplifyingSemiSupervised2020} & 64.3  & 49.4  & 47.3  & 30.9  & 18.5  & 15.3 & - & -  & -\\
    DS\textsuperscript{3}L \cite{guoSafeDeepSemiSupervised2020} & 70.5  & 46.6  & 43.5  & 33.7  & 15.8  & 15.1 & - & -  & -\\
    DTC \cite{han2019learning} & 42.7  & 31.8  & 32.4  & 22.1  & 10.5  & 13.7 & - & -  & - \\
    RankStats \cite{han2020automatically} & 71.4  & 63.9  & 66.7  & 20.4  & 16.7  & 17.8 & - & -  & -\\
    UNO \cite{fini2021unified} & 86.5  & 71.2  & 78.9  & 53.7  & 33.6  & 42.7 & 66.0 & 42.2  & 53.3 \\
    ORCA \cite{caoOpenWorldSemiSupervisedLearning2022} & 82.8  & 85.5  & 84.1  & 52.5  & 31.8  & 38.6 & 83.9 & 60.5  & 69.7 \\
    NACH \cite{guoRobustSemiSupervisedLearning2022} & 91.8  & 89.4  & 90.6  & 65.8  & 37.5  & 49.2 & - & -  & -  \\
    OpenLDN \cite{rizveOpenLDNLearningDiscover2022}  & 92.4  & 93.2  & 92.8  & 55.0  & 40.0  & 47.7 & - & -  & - \\
    TRSSL \cite{rizve2022realistic} & 94.9  & 89.6  & 92.2  & 68.5  & \textbf{52.1}  & \textbf{60.3} & 82.6 & 67.8  & 75.4 \\
    OpenCon \cite{sun2023opencon} & - & -  & - & 62.5 & 44.4  & 48.2 & - & -  & - \\
    \midrule
    OwMatch & 89.3  & 92.2  & 90.7  & 59.5  & 43.7  & 51.2 & 86.4 & 69.2 & 77.8 \\
    OwMatch+ & 94.4  & \textbf{96.2}  & \textbf{95.3}  & \textbf{69.9}  & 51.5  & \textbf{60.3}  & \textbf{87.8}  & \textbf{72.7}  & \textbf{80.2} \\
    \bottomrule
    \end{tabular}%
    \vskip -0.1 in
  \label{tab:main10}%
\end{table}%

\paragraph{Main results with label ratio of 10\%.} We evaluate our approach on CIFAR-10/100 and ImageNet-100, closely similar to Table~\ref{tab:main50}, but with the label ratio adjusted to 10\%. The results are detailed in Table~\ref{tab:main10}; OwMatch+ continues to maintain state-of-the-art performance across most benchmarks and evaluation metrics.

\begin{figure}[h]
    \centering
    \includegraphics[width=0.8\textwidth]{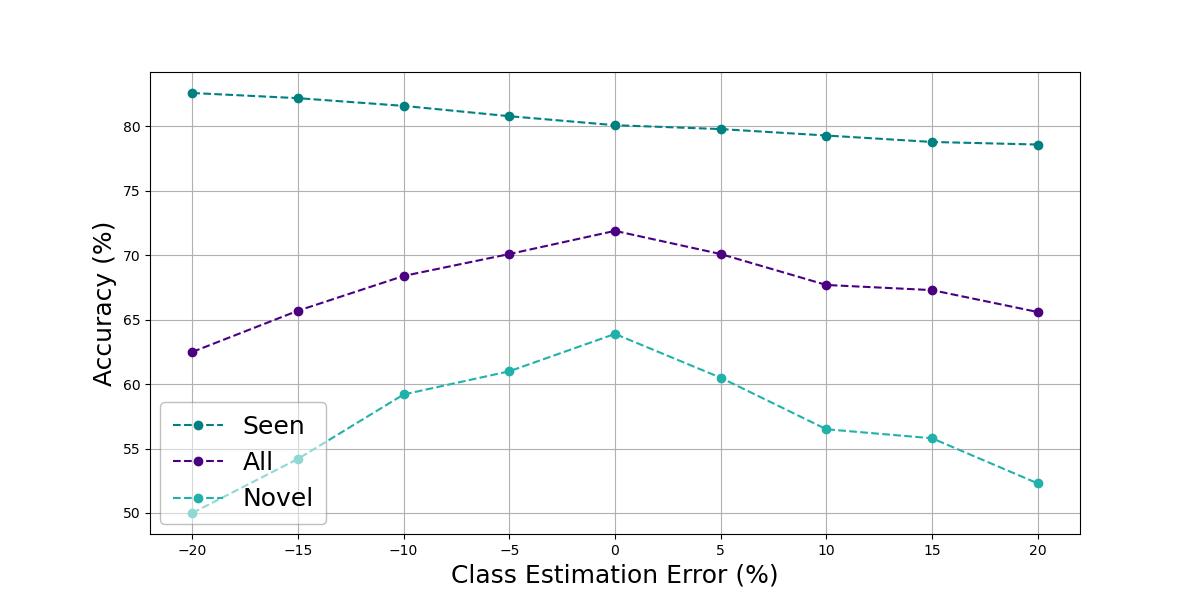}
    \caption{Accuracy as a function of class number estimation error on \textbf{CIFAR-100} dataset.}
    \label{fig_rebuttal:class_estimation_error}
\end{figure}

\paragraph{Performance sensitivity to varying extents of class number estimation error.} As illustrated in Figure \ref{fig_rebuttal:class_estimation_error}, OwMatch maintains robust performance over a range of errors.

\paragraph{Main results on Tiny-ImageNet.} In addition to the results on CIFAR and ImageNet datasets, we also evaluate OwMatch and OwMatch+ on Tiny-ImageNet with 50\% novel classes in Table~\ref{tab:tinyimage}. We found that both OwMatch and OwMatch+ surpass previous state-of-the-art methods across benchmarks and evaluation metrics.
\begin{table}[htbp]
    \footnotesize
    \centering
    \caption{Average accuracy on Tiny-ImageNet datasets with label ratio of 10\% and 50\%.}
 \begin{tabular}{@{}lccccccc@{}}
    \toprule
    \textbf{Method} & \multicolumn{3}{c}{\textbf{50\% label ratio}} & \multicolumn{3}{c}{\textbf{10\% label ratio}} \\
    \cmidrule(lr){2-4} \cmidrule(l){5-7} 
          & Seen  & Novel & All   & Seen  & Novel & All  \\
    \midrule
    DTC \cite{han2019learning} & 28.8  & 16.3  & 19.9  &  13.5 & 12.7  & 11.5  \\
    RankStats \cite{han2020automatically} & 5.7  & 5.4  & 3.4  &  9.6 & 8.9  & 6.4  \\
    UNO \cite{fini2021unified} & 46.5  & 15.7  & 30.3  &  28.4 & 14.4  & 20.4 \\
    OpenLDN \cite{rizveOpenLDNLearningDiscover2022} & 58.3 & 25.5  & 41.9  &  - & -  & -  \\
    TRSSL \cite{rizve2022realistic} & 59.1  & 24.2  & 41.7  &  39.5 & 20.5  & 30.3  \\
    \midrule
    OwMatch & 62.9  & 38.9  & 50.6 & 44.8 & 25.4  & 34.7  \\
    OwMatch+ & \textbf{68.8}  & \textbf{42.4}  & \textbf{55.0} & \textbf{55.8} & \textbf{33.3}  & \textbf{43.3}  \\
    \bottomrule
    \end{tabular}%
    \vskip -0.1 in
  \label{tab:tinyimage}%
\end{table}%

\paragraph{Different novel class ratio.} Previously, we assessed model's performance with a constant novel class ratio of 50\%, which is also variable in real-world scenarios. Here, we alter this value and fix the label ratio within seen classes to 50\%; the results are reported in Table~\ref{tab:abla_novelclassratio}. The model's performance generally exhibits a declining trend across all benchmarks and evaluation metrics as the novel class ratio increases. It is important to note that as the number of novel classes increases, the total amount of labeled data decreases.

\begin{table}[htbp]
    \footnotesize
    \centering
    \caption{Model performance across benchmarks with varying novel class ratios.}
 \begin{tabular}{@{}lcccccccccc@{}}
    \toprule
    \textbf{Novel class ratio} & \multicolumn{3}{c}{\textbf{CIFAR10}} & \multicolumn{3}{c}{\textbf{CIFAR100}} & \multicolumn{3}{c}{\textbf{Tiny-ImageNet}} \\
    \cmidrule(lr){2-4} \cmidrule(l){5-7} \cmidrule(l){8-10} 
          & Seen  & Novel & All & Seen & Novel & All & Seen & Novel & All \\
    \midrule
    50\% & 96.5  & 97.1 & 96.8 & 80.1  & 63.9 & 71.9 & 68.8  & 42.4 & 55.0\\
    60\% & 96.5  & 92.1  & 93.9 & 80.3  & 60.4 & 68.1 & 68.5  & 41.1 & 51.7 \\
    70\% & 97.5  & 91.0  & 93.0 & 81.3  & 58.6 & 65.3 & 71.3  & 34.7  & 45.6 \\
    80\% & 98.4  & 92.9  & 94.0  & 78.9  & 58.0 & 61.8 & 69.5  & 32.9  & 40.0 \\
    90\% & 97.8  & 93.6  & 94.0  & 82.0  & 50.7 & 53.5 & 69.4  & 26.4  & 30.3 \\
    \bottomrule
    \end{tabular}%
    \vskip -0.1 in
  \label{tab:abla_novelclassratio}%
\end{table}%

\begin{figure}[ht]
\begin{center}
\centerline{\includegraphics[width=0.65\columnwidth]{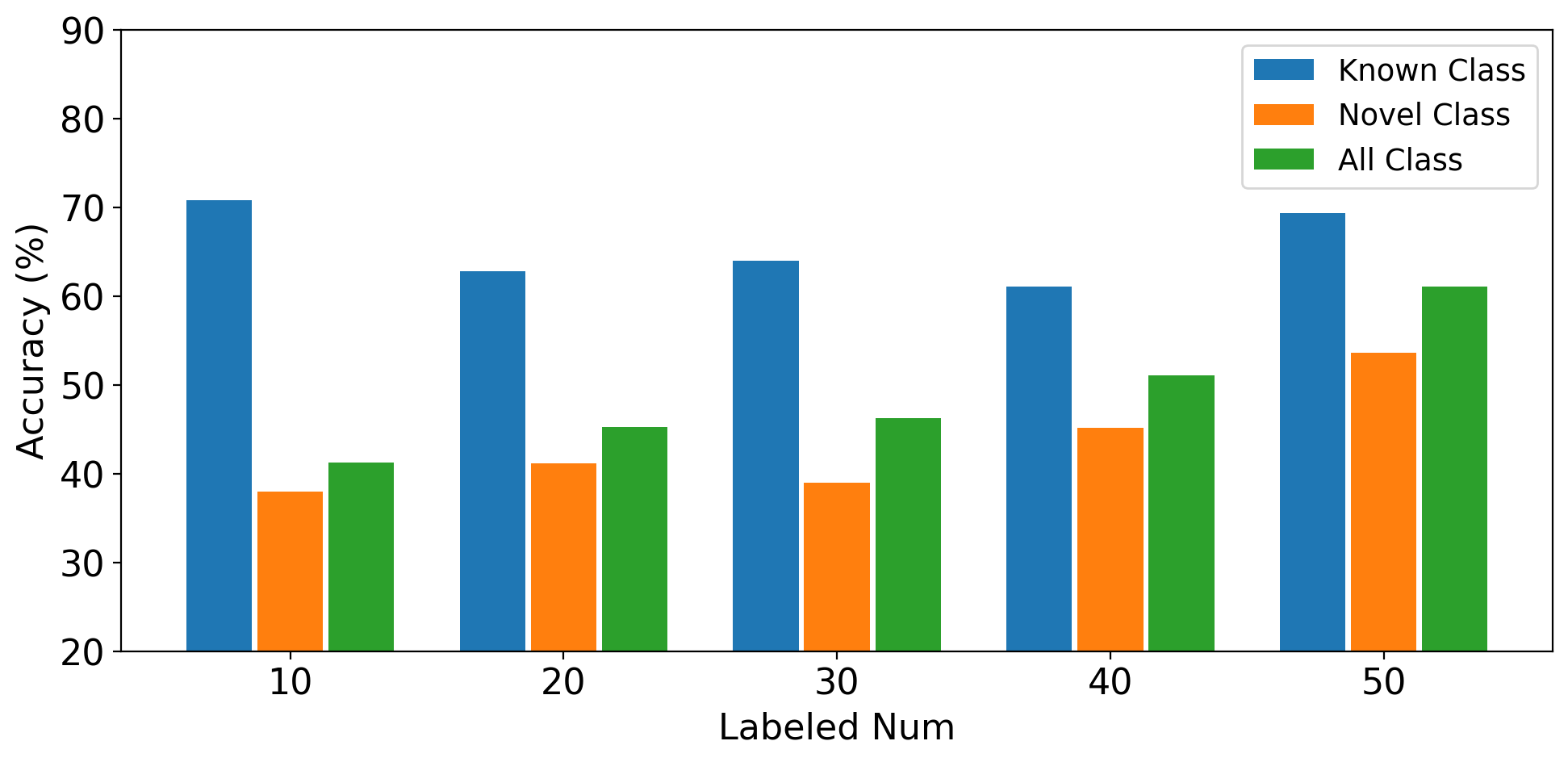}}
\caption{Model performance on CIFAR-100 with varying numbers of seen classes and a fixed amount of labeled data (5\% of total data). Models trained with more seen classes generally perform better on both novel and all classes, despite having fewer labeled data per seen class.}
\label{fig:supervision}
\end{center}
\end{figure}

\paragraph{Efficient labeling strategy under a fixed budget.}
In the previously conducted experiments, we assessed the model's performance by altering the novel class ratio and label ratio, respectively. Merely altering a single factor does not yield highly convincing inferences, as it is impractical to fix the novel class ratio or the label ratio for some classes in real-world scenarios. Here, we consider a fixed budget, specifically the total number of labeled data, as illustrated in Figure~\ref{fig:supervision}. This comparison aims to shed light on how the balance between labeled data in seen classes and the proportion of novel classes influences model performance under a fixed level of supervision.

When the supervisory information is held constant, a configuration with a smaller portion of labeled data spread across a greater number of different classes results in higher accuracy for both all classes and novel classes. Additionally, it is observed that as the number of novel classes decreases, the accuracy of the seen classes improves. This improvement is attributed to the reduced complexity of the classification task when there are fewer categories to be classified. From these observations, it can be inferred that within a limited labeling budget, it is more effective to label a broader category of samples, thereby capturing as many representative points as possible within the feature space. This strategy appears to optimize the model's performance across both known and novel classes.

\begin{figure}[htbp]
\centering
\begin{subfigure}[b]{0.42\textwidth}
    \includegraphics[width=\textwidth]{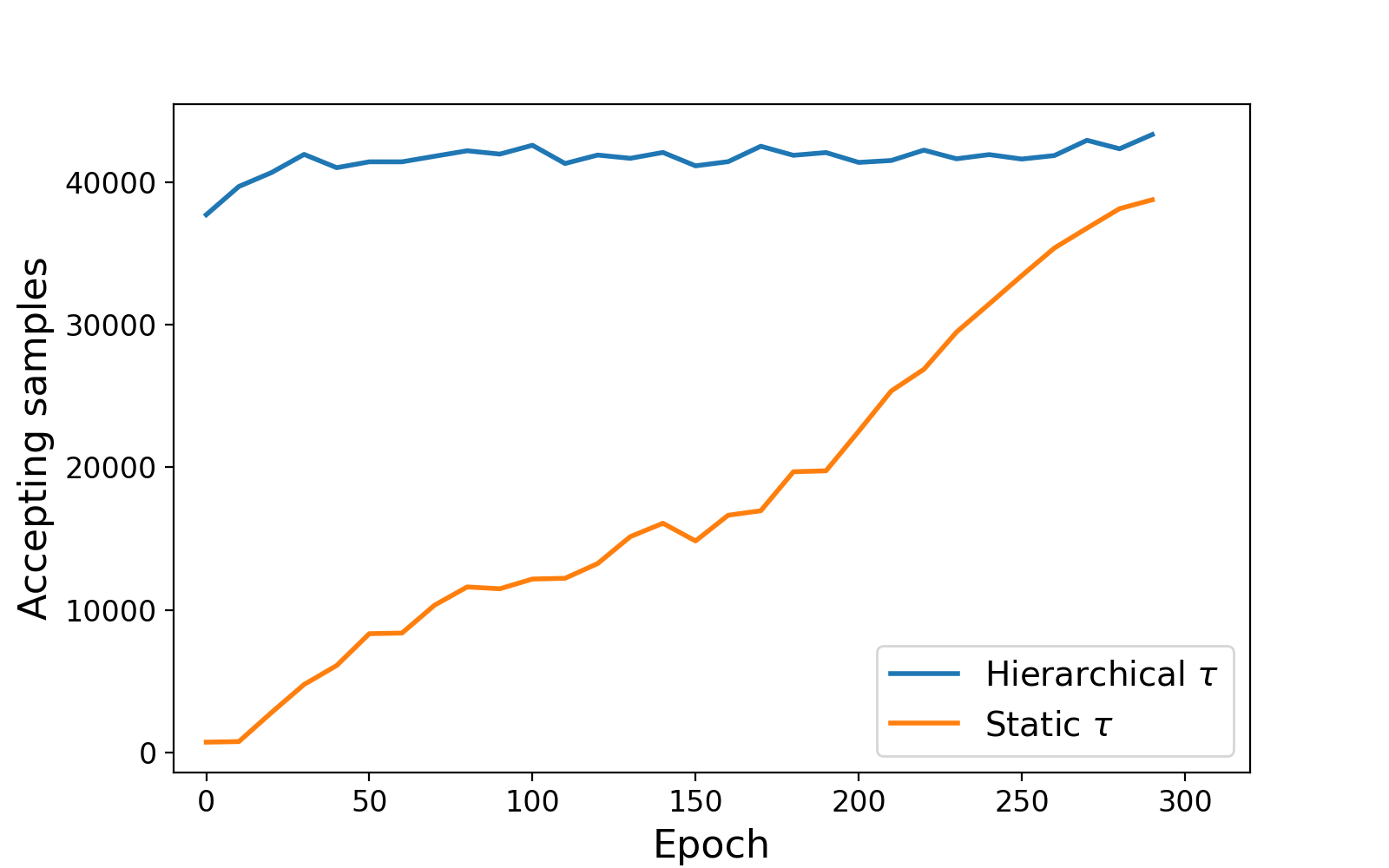}
    \caption{Number of selected pseudo-labels}
    \label{fig:ada_ablat_a}
\end{subfigure}
\begin{subfigure}[b]{0.42\textwidth}
    \includegraphics[width=\textwidth]{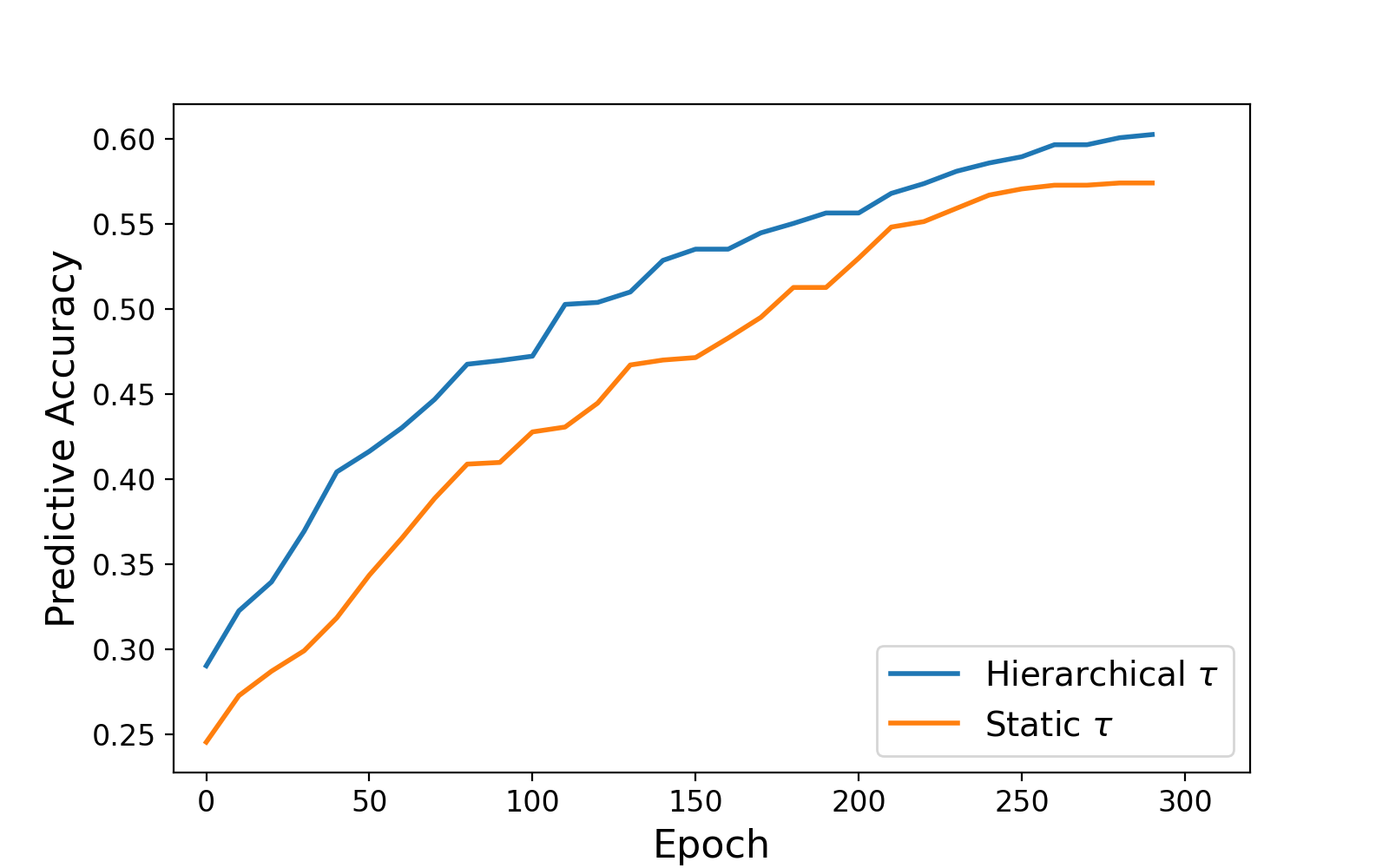}
    \caption{Pseudo-labels accuracy}
    \label{fig:ada_ablat_b}
\end{subfigure}%
\caption{Open-world hierarchical thresholding scheme generally selects more instances as pseudo-labels (a), and the quality of pseudo-labels is also enhanced (b).}
\label{fig:ada_ablat}
\end{figure}

\paragraph{Hierarchical thresholding scheme with scarce supervision information.}
Previously, we observed that in the scenario where the label ratio on seen classes is 50\%, the performance difference between adopting the hierarchical thresholding strategy and setting a high static threshold hyperparameter is not significant. Here, we maintain the novel class ratio as 50\% while decreasing the label ratio to 10\%.
Figure~\ref{fig:ada_ablat_a} and Figure~\ref{fig:ada_ablat_b} demonstrate that a hierarchical thresholding scheme not only retains more pseudo-labels but also preserves the predictive accuracy of selected pseudo-labels.

\begin{table}[htbp]
    \footnotesize
    \centering
    \caption{Model performance on CIFAR-10/100 and Tiny-ImageNet with different number of local views to contrastive learning.}
 \begin{tabular}{@{}lcccccccccc@{}}
    \toprule
    \textbf{\# of Crops} & \multicolumn{3}{c}{\textbf{CIFAR10}} & \multicolumn{3}{c}{\textbf{CIFAR100}} & \multicolumn{3}{c}{\textbf{Tiny-ImageNet}}\\
    \cmidrule(lr){2-4} \cmidrule(l){5-7} \cmidrule(l){8-10} 
          & Seen  & Novel & All & Seen  & Novel & All & Seen  & Novel & All \\
    \midrule
    2 & 95.9  & 97.7  & 96.8  &  78.8 & 61.6  & 70.2 & 66.4 & 43.6 & 54.2 \\
    4 & 96.5  & 97.1  & 96.8  &  80.1 & 63.9  & 71.9 & 68.8 & 42.4 & 55.0 \\
    6 & 96.8 & 97.2 & 97.0 &  81.1 & 59.4  & 69.8 & 70.1 & 42.1 & 55.6 \\
    \bottomrule
    \end{tabular}%
    \vskip -0.1 in
  \label{tab:crops}%
\end{table}%

\paragraph{Different number of local views in contrastive learning.} 
Contrastive learning boosts clustering by acquiring different views of the data and promoting consistency across these views at the representation level. The manner of data augmentation and the number of augmented views largely determine the model performance. Here we take the combination of color distortion and random resized crop \cite{chenSimpleFrameworkContrastive2020} as augmentation and apply a multi-crop technique \cite{caronUnsupervisedLearningVisual2020} to produce many low-resolution images (local views). We compare the performance with varying number of local views, the results are reported in Table~\ref{tab:crops}.

The incorporation of local views significantly enhances model performance. However, it's observed that as the number of local views increases, the incremental benefits to the model come to plateau, while the training and computational time considerably escalate. This aligns with the notion that local views assist in capturing local patterns, aiding in the development of robust representations. However, there is a threshold beyond which the addition of more local views contributes less to learning, overshadowed by the rise in computational demands.

\paragraph{Different number of iterations in the Sinkhorn-Knopp algorithm.} 
Conditional self-labeling is proposed to optimize high-quality self-label assignments, which depends on a fast version of the Sinkhorn-Knopp algorithm \cite{cuturi2013sinkhorn} to solve this complex linear programs efficiently. Resolving this involves iterative processes to converge on the optimal solution. We evaluate the model performance across different iterations, with the results presented in Table~\ref{tab:iteration}.

Generally, model performance tends to increase with more iterations of the Sinkhorn-Knopp algorithm. Despite the positive correlation tendency, we observe that certain iterations (e.g., 6 for CIFAR-100 with both novel class ratio and label ratio of 50\%) already achieve satisfactory outcomes, with only marginal gains from further iterations.

\begin{table}[htbp]
    \footnotesize
    \centering
    \caption{Model performance on CIFAR-10/100 and Tiny-ImageNet with different iteration numbers in Sinkhorn-Knopp algorithm.}
 \begin{tabular}{@{}lcccccccccc@{}}
    \toprule
    \textbf{SK-iters} & \multicolumn{3}{c}{\textbf{CIFAR10}} & \multicolumn{3}{c}{\textbf{CIFAR100}} & \multicolumn{3}{c}{\textbf{Tiny-ImageNet}} \\
    \cmidrule(lr){2-4} \cmidrule(l){5-7} \cmidrule(l){8-10}
          & Seen  & Novel & All   & Seen  & Novel & All & Seen  & Novel & All \\
    \midrule
    3 & 96.8  & 91.6  & 94.2  &  81.5 & 55.0  & 67.6 & 68.6 & 41.7 & 54.1 \\
    6 & 96.7  & 91.0  & 93.9  &  81.4 & 62.4  & 71.3 & 68.0 & 42.8 & 54.5 \\
    10 & 96.5  & 97.1  & 96.8 &  80.1 & 63.9  & 71.9 & 68.8 & 42.4 & 55.0 \\
    \bottomrule
    \end{tabular}%
    \vskip -0.1 in
  \label{tab:iteration}%
\end{table}%

\section{Proof details}
\label{app:proof}

Here, we derive ECS for both unconditional and conditional self-labeling in the following two lemmas.

\begin{lemma}
\label{thm:uncon}
Estimators of $\mathbf{A}$ from unconditional self-labeling has the form of
\begin{equation}
    \mathbf{A}_{\mathrm{uncon}} = [N^up_1,N^up_2,\cdots,N^up_K],
\end{equation}
and ECS for $\hat{\boldsymbol{\mu}}_{\mathrm{uncon}} = \frac{1}{N^u} \mathbf{A}_{\mathrm{uncon}}$ can be derived as
\begin{align}
    \mathrm{ECS}(\hat{\boldsymbol{\mu}}_{\mathrm{uncon}})= \sum_{i=1}^{K} \frac{N^u(p_i-p_i^u)^2}{p_i^u}. 
\end{align}
\end{lemma}

\begin{lemma}
\label{thm:consela}
Estimators of $\mathbf{A}$ from conditional self-labeling has the form of
\begin{align}
    \mathbf{A}_{\mathrm{con}}=[Np_1-N^l_1,\cdots, Np_{K}-N^l_{K}],
\end{align}
and ECS for $\hat{\boldsymbol{\mu}}_{\mathrm{con}}=  \frac{1}{N^u} \mathbf{A}_{\mathrm{con}}$ can be derived as
\begin{align}
     \mathrm{ECS}(\hat{\boldsymbol{\mu}}_{\mathrm{con}})=\sum_{i=1}^{K} \frac{N^l p^l_i(1-p^l_i)}{N^u p_i^u}. 
\end{align}
\end{lemma}

\subsection{Proof of Lemma~\ref{lemma:chi}}
\begin{proof}
Under the null hypothesis $H_0$, the sample size of the $i$-th class follows the Binary distribution with parameters of $N$ and $p_i$. Therefore, we have the expectation and standard error of $N_i$ as
$\mathbb{E}_{\boldsymbol{\mathcal{P}}}(N_i) = Np_i$ and $\mathbb{SD}(N_i) = Np_i(1-p_i)$, respectively, where the standard deviation of $N_i$ measures the average deviation of random variable $N_i$ from its expected value. And for observation $n_i$ for each class, the discrepancy can be denoted as $n_i - \mathbb{E}_{\boldsymbol{\mathcal{P}}}(N_i)$. To ensure discrepancies for each class are evaluated on a consistent basis, dividing them by their standard errors under $H_0$, which enables us to focus on the standardized variable $\frac{n_i - \mathbb{E}_{\boldsymbol{\mathcal{P}}}(N_i)}{(Np_i(1-p_i))^{1/2}}$.

Note that $\sum_{i=1}^{K} \mathbb{E}_{\boldsymbol{\mathcal{P}}}(N_i) = \sum_{i=1}^{K} Np_i = N$, therefore the discrepancies across the $K$ classes cannot simultaneously be all positive or all negative. To measure the discrepancies of all $K$ classes, we sum the squares of the discrepancies of each class. This formulation ensures that discrepancies are independent of sign, merely reflecting the deviation between observed and expected values under $H_0$:
\begin{equation}
    \frac{(n_1-Np_1)^2}{Np_1(1-p_1)} + \frac{(n_2-Np_2)^2}{Np_2(1-p_2)} + \cdots + \frac{(n_{K}-Np_{K})^2}{Np_i(1-p_{K})}.
\end{equation}
We then prefer to exclude the factors $(1-p_i)$ from the denominators of the sum for two primary reasons.  Firstly, if numerous classes exist and none of them has significant large probabilities, then $(1-p_i)$ is almost negligible. Moreover, when the expectation $\mathbb{E}_{\boldsymbol{\mathcal{P}}}(N_i)$ is substantial, the approximation of each discrepancy adheres to a standard normal distribution, which leads to the construct of chi-square statistics with the degree of freedom of $K-1$. Then the summary statistic can be obtained as follows
\begin{equation}
    \chi^2 = \sum_{i=1}^{K} \frac{(n_i-Np_i)^2}{Np_i} = \sum_{i=1}^{K}\frac{(n_i-\mathbb{E}_{\boldsymbol{\mathcal{P}}}[N_i]^2)}{\mathbb{E}_{\boldsymbol{\mathcal{P}}}[N_i]}\sim \chi^2_{K-1}.
\end{equation}
\end{proof}

\subsection{Proof of Lemma~\ref{thm:uncon}}
\label{app:proof_uncon}
\begin{proof}
TRSSL \cite{rizve2022realistic} posits that unlabeled and real data share the same class distribution, and optimize the self-label assignment on unlabeled data solely based on prior information $\boldsymbol{\mathcal{P}}$. Owing to the alignment constraint between the generative self-label assignment and prior information, as shown in \eqref{eq:res_uncon}. Then the estimated number of samples in each class from conditional self-labeling is
\begin{equation}
    \mathbf{A}_{\mathrm{uncon}} = [N^up_1,N^up_2,\cdots,N^up_{K}],
\end{equation}
where both $N^u$ and $p_1,\cdots, p_{K}$ are known, therefore $\mathbf{A}_{\mathrm{uncon}}$ is a constant vectors. Then the estimator $\hat{\boldsymbol{\mu}}_{\mathrm{uncon}} = \frac{1}{N^u} \mathbf{A}_{\mathrm{uncon}}$ is also a constant vectors. Thus, the ECS for $\hat{\boldsymbol{\mu}}_{\mathrm{uncon}}$ can be calculated directly,
\begin{align}
    \mathrm{ECS}(\hat{\boldsymbol{\mu}}_{\mathrm{uncon}})= &\mathbb{E}\left[ \sum_{i=1}^{K} \frac{(N^up_i-\mathbb{E}_{\boldsymbol{\mathcal{P}}}[N^u_i])^2}{\mathbb{E}_{\boldsymbol{\mathcal{P}}}[N^u_i]}\right] \\
    = & \sum_{i=1}^{K} \frac{(N^up_i-N^up_i^u))^2}{N^u p_i^u} \\
    = &\sum_{i=1}^{K} \frac{N^u(p_i-p_i^u)^2}{p_i^u}.
\end{align}
\end{proof}

\subsection{Proof of Lemma~\ref{thm:consela}}
\label{app:proof_con}
\begin{proof}
Different from an unconditional setting, conditional self-labeling considers partial supervision. Specifically, except for the constraint brought from prior information, an additional constraint is constructed to realize the alignment between clustering results and existing labels in labeled data, as shown in \eqref{eq:res_con}. Then the estimated number of samples in each class from conditional self-labeling is
\begin{align}
    \mathbf{A}_{\mathrm{con}}=[Np_1-N^l_1,\cdots, Np_{K}-N^l_{K}],
\end{align}
where $N, p_1,p_2,\cdots, p_{K}$ are static values, and $N^l_1,\cdots, N^l_{K}$ are a set of random variables. Note that in a specific experiment, we can get a set of observations of $N^l_1,\cdots, N^l_{K}$, thus estimation based on conditional self-labeling is feasible. Then consider the estimator $\hat{\boldsymbol{\mu}}_{\mathrm{con}} = \frac{1}{N^u} \mathbf{A}_{\mathrm{con}}$, derivation steps of the ECS for $\hat{\boldsymbol{\mu}}_{\mathrm{con}}$ are as follow,
\begin{align}
     \mathrm{ECS}(\hat{\boldsymbol{\mu}}_{\mathrm{uncon}})= \mathbb{E}\left[ \sum_{i=1}^{K} \frac{(Np_i-N^l_i-\mathbb{E}_{\boldsymbol{\mathcal{P}}}[N^u_i])^2}{\mathbb{E}_{\boldsymbol{\mathcal{P}}}[N^u_i]}\right] = \mathbb{E}\left[ \sum_{i=1}^{K} \frac{(Np_i-N^l_i- N^up^u_i)^2}{N^up^u_i}\right].
\end{align}
According to  \eqref{eq:basic},  we have
\begin{align}
     \mathbb{E}\left[ \sum_{i=1}^{K} \frac{(Np_i-N^l_i- N^up^u_i)^2}{N^up^u_i}\right]
    = \mathbb{E}\left[ \sum_{i=1}^{K} \frac{(N^l_ip^l_i-N^l_i)^2}{N^up^u_i}\right]
    = \mathbb{E}\left[ \sum_{i=1}^{K} \frac{(N^l_i - \mathbb{E}_{\boldsymbol{\mathcal{P}}}[N^l_i])^2}{N^up^u_i}\right].
\end{align}
Since ECS is also defined at the population level, thus we have
\begin{align}
     \mathbb{E}\left[ \sum_{i=1}^{K} \frac{(N^l_i - \mathbb{E}_{\boldsymbol{\mathcal{P}}}[N^l_i])^2}{N^up^u_i}\right]
     = \left[ \sum_{i=1}^{K} \frac{\mathbb{E} (N^l_i - \mathbb{E}_{\boldsymbol{\mathcal{P}}}[N^l_i])^2}{N^up^u_i}\right]
     = \sum_{i=1}^{K} \frac{\text{Var}(N^l_i)}{N^up^u_i}. 
\end{align}
Since $N^l_1,N^l_2,\cdots,N^l_{K} \sim \text{Multinomial}(N^u,\boldsymbol{\mathcal{P}}^l)$, we have
\begin{align}
     \sum_{i=1}^{K} \frac{\text{Var}(N^l_i)}{N^up^u_i}
     = \sum_{i=1}^{K} \frac{N^l p^l_i (1-p^l_i)}{N^up^u_i}.
\end{align}
\end{proof}

\subsection{Proof of Theorem~\ref{thm:unbias}}
\label{app:proof_unbias}
\begin{proof}
From Theorem~\ref{app:proof_uncon} and Theorem~\ref{app:proof_con}, we have that
\begin{align}
    &\hat{\boldsymbol{\mu}}_{\mathrm{uncon}} = [p_1,p_2,\cdots, p_{K}] = \boldsymbol{\mathcal{P}}, \\
    &\hat{\boldsymbol{\mu}}_{\mathrm{con}} =\left[\frac{Np_1-N^l_1}{N^u},\cdots, \frac{Np_{K}-N^l_{K}}{N^u}\right]. 
\end{align}
Note that $\hat{\boldsymbol{\mu}}_{\mathrm{uncon}}$ is exactly the prior information, $\hat{\boldsymbol{\mu}}_{\mathrm{con}}$ are function of a set of random variables $N^l_i,N^l_2,\cdots, N^l_{K}$, then we have
\begin{align}
    \mathbb{E}(\hat{\boldsymbol{\mu}}_{\mathrm{con}}) & =\left[\mathbb{E}(\frac{Np_1-N^l_1}{N^u}),\cdots, \mathbb{E}(\frac{Np_{K}-N^l_{K}}{N^u})\right] \\ & = \left[\frac{Np_1-N^lp^l_1}{N^u}, \cdots, \frac{Np_1-N^lp^l_{K}}{N^u}\right]. 
\end{align}
According to \eqref{eq:basic}, we have
\begin{align}
    \left[\frac{Np_1-N^lp^l_1}{N^u}, \cdots, \frac{Np_1-N^lp^l_{K}}{N^u}\right] & = \left[\frac{N^up^u_1}{N^u}, \cdots, \frac{N^up^u_{K}}{N^u}\right] \\ & = \left[p_1^u,p_2^u,\cdots, p_{K}^u\right]= \boldsymbol{\mathcal{P}}^u.
\end{align}
Thus, $\hat{\boldsymbol{\mu}}_{\mathrm{con}}$ is a unbiased estimator of $\mathcal{C}^u$.
\end{proof}

\subsection{Proof of Theorem~\ref{thm:variablity}}
\begin{proof}
    Note that $r_i := \frac{N^l \cdot p_i^l}{N}$ is non-negative and obtain zero if and only if the $i$-th class denote a novel class.\\
    When the $i$-th class refers to seen classes, we have $p^u_i = \frac{p_i-r_i}{1-{r}}$, then
    \begin{align*}
        N^u(p_i-p_i^u)^2 \geq N^u(r_i-r \cdot p_i^u)^2 \geq 1.
    \end{align*}
    When the $i$-th class refers to an novel class, we have $p^u_i = \frac{p_i}{1-r}$, then
    \begin{align*}
        N^u(p_i-p_i^u)^2 = N^u\cdot \left(p_i-\frac{p_i}{1-r}\right)^2 = N^u\cdot \left(\frac{rp_i}{1-r}\right)^2 \geq N^u (r\cdot p_i)^2 \geq 1.
    \end{align*}
    Therefore, we get that 
    \begin{align}
        \mathrm{ECS}(\hat{\boldsymbol{\mu}}_{\mathrm{uncon}})-\mathrm{ECS}(\hat{\boldsymbol{\mu}}_{\mathrm{con}}) = & \sum_{i=1}^{K}\frac{N^u(p_i-p^u_i)^2-\frac{N^l}{N^u}p^l_i(1-p^l_i)}{p_i^u} \\
        \geq &\sum_{i=1}^{K}\frac{N^u(p_i-p^u_i)^2-1}{p_i^u}
        \geq  0.
    \end{align}
\end{proof}

\section{Broader impacts}
\label{app:impact}
This research delves into the issue of semi-supervised learning (SSL) in situations where not all classes possess labeled instances, an aspect that has received limited attention within the realm of SSL. We aim to draw increased focus towards examining the resilience of SSL in diverse real-world scenarios, thereby fostering a broader application of SSL in various contexts. However, the current accuracy is not very high for some challenging datasets. Therefore, the predictive results should be best used as references rather than treated as ground truth.

\section{Limitations}
\label{app:limit}
OwMatch, similar to existing methods in OwSSL, faces a significant challenge when applied to imbalanced datasets or unknown prior class distribution. Existing OwSSL methods are typically applied on class-balanced datasets where instances of each class share nearly the same frequency; the model performance would deteriorate when encountering imbalanced datasets. On the other hand, prior class distributions are not available in real-world applications. Addressing the dependency on prior class distribution and effectively handling datasets of arbitrary composition remain challenging for existing OwSSL algorithms, including OwMatch.

Recognizing this, we propose an adaptive estimation scheme for the OwMatch framework and show its feasibility in the experiments, with results reported in Table~\ref{tab:abla_imb}. Although a performance decline within 3\% may seem acceptable, it is worth further consideration and exploration to determine whether further optimizations can enhance model performance without any prior. At the same time, we aim to prove the convergence of this adaptive algorithm in our future work.


\end{document}